\documentclass[10pt,twocolumn,letterpaper]{article}

\usepackage{iccv}              

\usepackage{kotex}
\usepackage{multirow}
\usepackage{colortbl}
\usepackage{xcolor}
\usepackage{color}

\usepackage{booktabs}
\usepackage{bbding}
\usepackage{amsmath}
\usepackage{amssymb}
\usepackage{xparse}
\usepackage{makecell}
\usepackage{arydshln}

\definecolor{orange}{HTML}{F6E3CE}

\definecolor{myred}{RGB}{204, 51, 51}   
\definecolor{myblue}{RGB}{51, 102, 204} 

\definecolor{iccvblue}{rgb}{0.21,0.49,0.74}
\usepackage[pagebackref,breaklinks,colorlinks,allcolors=iccvblue]{hyperref}

\def\paperID{11796} 
\def\confName{ICCV}
\def\confYear{2025}

\title{Emulating Self-attention with Convolution for Efficient Image Super-Resolution}

\author{Dongheon Lee, \qquad Seokju Yun, \qquad Youngmin Ro\textsuperscript{*}\\
Machine Intelligence Laboratory, University of Seoul, Korea\\
\small{\texttt{\{dslisleedh, wsz871, youngmin.ro\}@uos.ac.kr}} \\
  \small{Code: \href{https://github.com/dslisleedh/ESC}{https://github.com/dslisleedh/ESC}}
}

\begin{document}
\maketitle
\renewcommand{\thefootnote}{\fnsymbol{footnote}}
\footnotetext[1]{Corresponding author.}
\vspace{-0.5cm}
\begin{abstract}
In this paper, we tackle the high computational overhead of Transformers for efficient image super-resolution~(SR).
Motivated by the observations of self-attention's inter-layer repetition, we introduce a convolutionized self-attention module named Convolutional Attention~(ConvAttn) that emulates self-attention's long-range modeling capability and instance-dependent weighting with a single shared large kernel and dynamic kernels.
By utilizing the ConvAttn module, we significantly reduce the reliance on self-attention and its involved memory-bound operations while maintaining the representational capability of Transformers.
Furthermore, we overcome the challenge of integrating flash attention into the lightweight SR regime, effectively mitigating self-attention's inherent memory bottleneck.
We scale up the window size to 32$\times$32 with flash attention rather than proposing an intricate self-attention module, significantly improving PSNR by 0.31dB on Urban100$\times$2 while reducing latency and memory usage by 16$\times$ and 12.2$\times$.
Building on these approaches, our proposed network, termed Emulating Self-attention with Convolution~(ESC), notably improves PSNR by 0.27 dB on Urban100$\times$4 compared to HiT-SRF, reducing the latency and memory usage by 3.7$\times$ and 6.2$\times$, respectively.
Extensive experiments demonstrate that our ESC maintains the ability for long-range modeling, data scalability, and the representational power of Transformers despite most self-attention being replaced by the ConvAttn module.
\end{abstract}

\vspace{-0.6cm}
\section{Introduction}
\vspace{-0.1cm}
Image Super-Resolution~(SR) aims to reconstruct high-resolution (HR) images from low-resolution~(LR) inputs and remains an active area of research in computer vision.
Recently, the demand for multimedia content and generative models has increased significantly, making SR particularly noteworthy as it enables users to enjoy high-quality content under resource-constrained conditions.
Consequently, practical deployment has emerged as a critical consideration in SR tasks, motivating numerous SR studies to enhance performance while reducing computational complexity~(FLoating point OPerations; FLOPs) and parameter size.

Therefore, Transformers have gained significant attention in the SR task since they have achieved superior performance over Convolutional Neural Networks~(CNNs) while requiring lower FLOPs and fewer parameters.
By capturing long-range dependencies and performing input-dependent weighting through self-attention, Transformers exhibit high representational capacity with enhanced performance, especially as the training data volume increases~\cite{ViT, IPT}.
Accordingly, numerous transformer-based methods have been proposed~\cite{ELAN, OmniSR, ASID} to capitalize on the advantages of self-attention while reducing FLOPs and parameter counts.

\begin{figure}[t]
  \centering
  \includegraphics[width=\columnwidth]{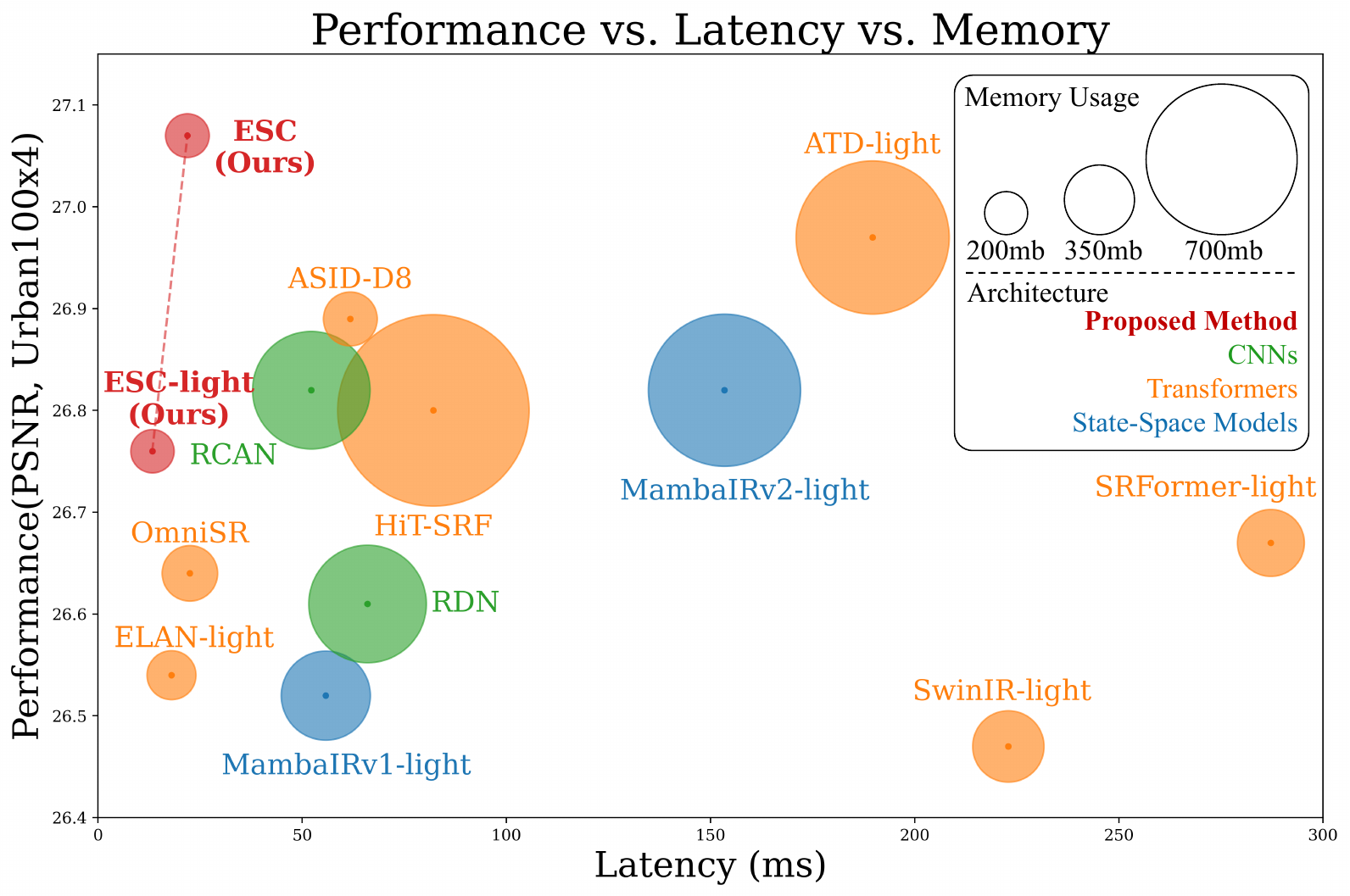}
  \vspace{-0.6cm}
  \caption{
    Comparison of performance, latency, and memory usage. 
    Our methods are evaluated against representative SR models, including CNNs, Transformers, and SSMs.
  }
  \vspace{-0.6cm}
  \label{fig:teaser}
\end{figure}

However, many of these works overlook the excessive memory access caused by self-attention, which arises from materializing the score matrix~($S=QK^{T}$)~\cite{FlashAttn} and leveraging memory-bound operations such as tensor reshaping and window masking~\cite{SHViT, DITN}.
The memory access issue is further worsened in the SR architecture, which deals with large feature maps without patchify stem or downsampling stage.
For example, SwinIR-light~\cite{SwinIR} is 4.7$\times$ slower latency and 2$\times$ higher memory usage than CNNs~\cite{RCAN} reconstructing an HD image at the $\times$2 scale, even though it requires 14.5$\times$ and 17$\times$ fewer FLOPs and parameter sizes, respectively.
Consequently, despite their promising performance, Transformers are challenging to deploy on resource-constrained devices, such as consumer-level GPUs.

\begin{figure}[t]
  \centering
  \includegraphics[width=\columnwidth]{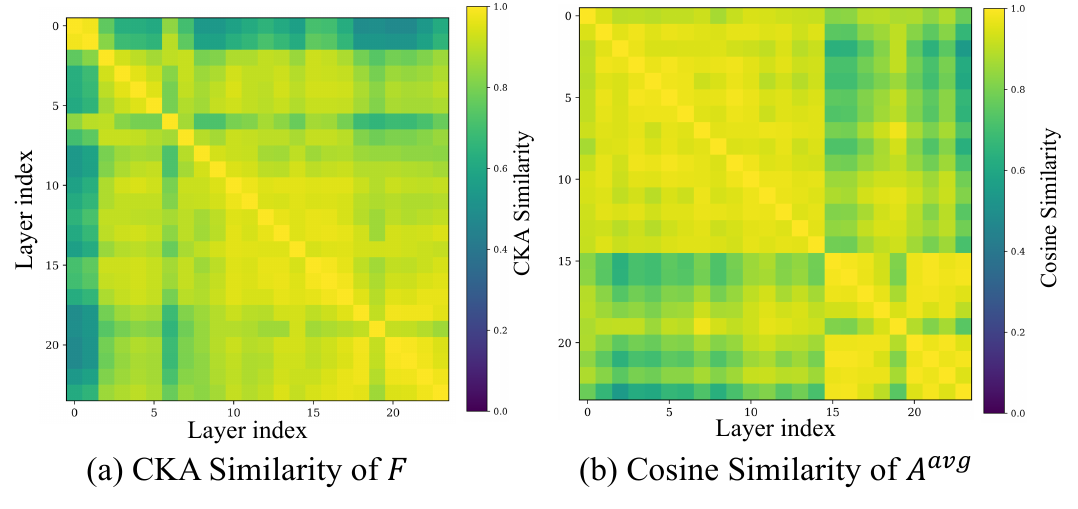}
  \vspace{-0.6cm}
  \caption{
    Visualized inter-layer similarities of the feature extracted by self-attention~($F$) before being added to the skip path and corresponding attention map~($A^{avg}$) from SwinIR-light~\cite{SwinIR}.
    Both CKA~\cite{CKA} and cosine similarities are measured on the Urban100$\times$4 dataset.
    The window and head dimensions of the attention map are averaged for visualization.
  }
  \vspace{-0.6cm}
  \label{fig:cka}
\end{figure}

In this paper, we aim to design an efficiency-driven Transformer tailored for lightweight SR tasks, achieving both improved performance and memory overhead.
We start by conducting a preliminary analysis of self-attention, observing that the similarity modeling performed by self-attention and resulting extracted features remain highly consistent across several layers, as shown in Figure~\ref{fig:cka}.
This finding exhibits that self-attention may extract overlapping features, suggesting the possibility of reducing computational overhead without compromising representational capability by utilizing efficient alternatives.
Building on this finding, we establish a design strategy that retains self-attention only in the first layer of each block while replacing the remaining layers with our proposed efficient alternative, the Convolutional Attention~(ConvAttn) module. 
To effectively emulate self-attention's long-range modeling and instance-dependent weighting, the ConvAttn module operates with a twofold mechanism. 
First, it simplifies self-attention's long-range interaction by applying convolution with a shared 13$\times$13 large kernel throughout the entire network, targeting only a subset of channels.
Second, dynamic kernels are generated to capture input-dependent weighting, mimicking the adaptive nature of self-attention. 
By combining these components, the ConvAttn module significantly reduces the reliance on memory-intensive self-attention while maintaining the representational capability of Transformers. 

With most self-attention layers replaced by ConvAttn, we leverage this efficiency to further enhance the remaining self-attention layers. 
Specifically, we enlarge the window size of self-attention, significantly improving performance with only a slight increase in FLOPs.
However, increasing the window size leads to an enlarged score matrix, substantially raising peak memory usage.
To address this, we introduce Flash Attention~\cite{FlashAttn, FlexAttn} into lightweight SR tasks to avoid materializing the score matrix.
Our optimized implementation allows us to scale up the window size to 32$\times$32 while reducing latency and memory usage by 16$\times$ and 12.2$\times$, respectively, as illustrated in Table~\ref{tab:attn_compare}.

Based on these approaches, we introduce the lightweight SR network called Emulating Self-attention with Convolution~(ESC).
The proposed ESC outperforms ATD-light~\cite{ATD} on Urban100$\times$4 with 0.1dB improvements on PSNR while being 8.9$\times$ faster, as shown in Figure~\ref{fig:teaser}.
Additionally, ESC-light surpasses ELAN-light~\cite{ELAN} by a large margin of 0.29dB in PSNR on Urban100$\times$2 while also reducing latency by 22\%.
We further validate our ESC in scenarios where reducing FLOPs and parameter size is essential by introducing ESC-FP, which outperforms MambaIRV2-light~\cite{MambaIRV2} on Manga109$\times$4 with reductions of FLOPs and parameter size by 20\% and 32\%, respectively.
Through our extensive experiments, we demonstrate that our ESC fully leverages the advantages of Transformers -- including their large receptive fields, representational capacity, and scalability with respect to data volume -- even though most self-attentions are replaced by the ConvAttn module.
We support these results with in-depth experiments, as illustrated in Figure~\ref{fig:frequency}, suggesting that the proposed ConvAttn module extracts similar features with self-attention.

Our contributions are summarized as follows:
\begin{itemize}
    \item We demonstrate that well-designed convolution partially replaces self-attention, significantly improving efficiency without sacrificing the advantages of Transformers.
    \item We mitigate the memory overhead of self-attention by integrating Flash Attention, thereby enlarging the window size to 32$\times$32 without excessive memory usage. To the best of our knowledge, this is the first successful application of Flash Attention in lightweight SR tasks.
    \item We fully exploit the transformer's benefits in lightweight SR tasks by making them more efficient and simpler.
\end{itemize}

\vspace{-0.15cm}
\section{Related Work}
\noindent\textbf{Traditional CNNs}
Early deep learning‐based SR research~\cite{SRCNN, VDSR, EDSR, RDN, RCAN} primarily employed convolution that extracts local features from images.
These approaches typically relied on stacked convolutions, which often resulted in over‐parameterization. 
To address this issue, numerous studies~\cite{DRCN, DRRN, SRFBN, PFS} have shared their weights across multiple layers to reduce parameter size.
However, many of them suffer from performance degradation since the input-independent weights are shared across multiple layers.
In this paper, we share a large kernel through the entire network while leveraging input-dependently generated dynamic Depth-Wise Convolution~(DWC)~\cite{DWNet} by each layer in parallel.
This approach efficiently mitigates the parameter growth of the large kernel while maintaining each layer's representational power.  

\noindent\textbf{Transformers}
Pioneering studies that significantly reduce computational costs by computing self-attention within local windows~\cite{SwinIR} or channel-wise~\cite{Restormer} have sparked considerable interest in applying Transformers~\cite{Transformers, ViT} to the SR task.
Subsequent studies have proposed methods that enhance self‐attention to strengthen its long‐range modeling capabilities while reducing both FLOPs and parameters~\cite{ELAN, OmniSR, HAT, SRFormer, ATD, HiTSR, ASID}. 
Nonetheless, Transformers remain challenging to deploy due to the excessive memory access caused by self-attention and involved memory-bound operations. 
In this paper, we address this limitation by replacing self‐attention with convolutions that emulate its advantages, based on the observations of self-attention's inter-layer repetition.
Our approach preserves the advantages of Transformers while reducing reliance on self-attention and associated memory-bound operations.

\noindent\textbf{Large Kernel CNNs and State-Space Models}
Several recent studies have highlighted the computational burden of Transformers and investigated alternative core operators to replicate their advantages. 
Specifically, CNN-based models~\cite{ShuffleMixer, VapSR, LKDN, ConvFormer} attempt to mimic the advantages of Transformers by leveraging large kernels and pixel attention~\cite{PAN}, while State-Space Models~(SSM)~\cite{MambaIR, MambaIRV2} adapt their verified long-range modeling abilities from the 1D sequence domain~\cite{S4, Mamba} to the 2D image domain.
However, these approaches often overlook either the exceptional representational capability brought by self-attention’s sophisticated similarity modeling or the additional memory access caused by supplementary mechanisms to adapting 2D domains like multi-directional or attentive scanning.
In this paper, we fully leverage the advantages of both self-attention and convolution while utilizing Flash Attention~\cite{FlashAttn, FlexAttn} to address the memory access of self-attention without relying on complex additional mechanisms.

\vspace{-0.15cm}
\section{Proposed Methods}
\vspace{-0.1cm}
In this section, we start by representing our preliminary analysis and then describe our proposed network.

\subsection{Preliminary analysis}
\vspace{-0.1cm}

Transformers often achieve improved performance by utilizing self-attention's long-range modeling ability and instance-dependent weighting.
However, each self-attention incurs excessive memory overhead, resulting in significant latency and memory usage.
Interestingly, our preliminary analysis demonstrates that the sophisticated similarity modeling of self-attention~($A^{avg}$) and the resulting extracted features ($F$) exhibit a high degree of inter-layer similarities, with averages of 89\% and 87\%.
This observation not only aligns with recent research~\cite{UPS} but also explains the success of studies that reduced computational cost by sharing the attention map across layers~\cite{ELAN, ASID}. 
Inspired by this observation, we investigate improving efficiency without compromising representational power.

\subsection{Overall Structure}
\begin{figure}[t]
  \centering
  \includegraphics[width=\columnwidth]{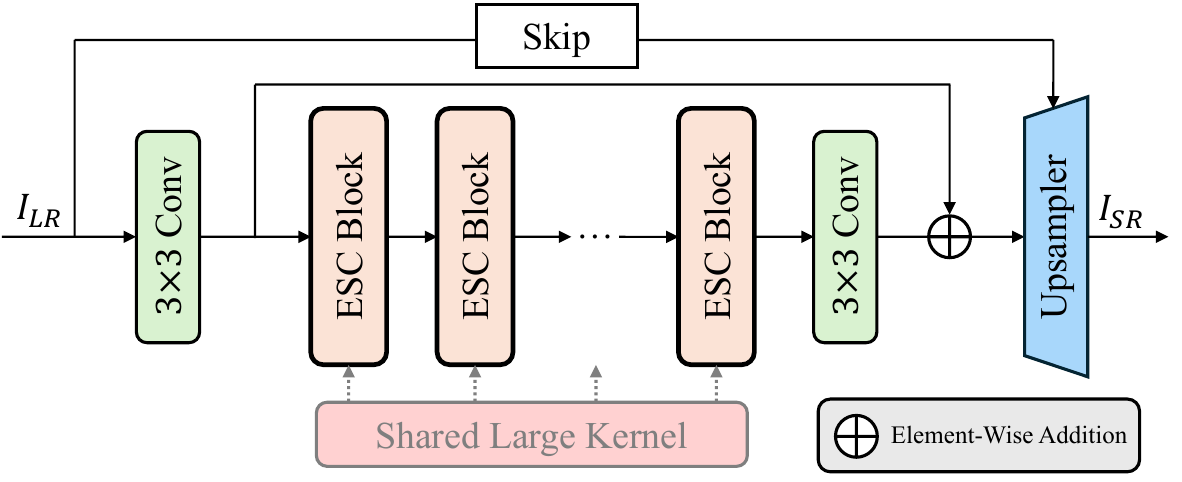}
  \vspace{-0.6cm}
  \caption{
    Visualization of the overall architecture.
  }
  \vspace{-0.5cm}
  \label{fig:overall_architecture}
\end{figure}

Our network is composed of four main components, as shown in Figure~\ref{fig:overall_architecture}.
First, the LR input image~$I_{LR}\in\mathbb{R}^{H\times W\times 3}$ is transformed into a shallow feature~$F_{0}\in\mathbb{R}^{H\times W\times C}$ through a 3$\times$3 convolution~($\mathrm{Conv}_{3\times3}$).
Here, $H$ and $W$ denote height and width of $I_{LR}$, and $C$ denotes feature dimension.
Next, $F_{0}$ and a large kernel~$LK\in\mathbb{R}^{13\times13\times16\times16}$ that is learned end-to-end manner and shared across the entire model are fed into the deep feature extractor $H$, and the resulting deep feature is added back to $F_{0}$ to produce the final feature $F\in\mathbb{R}^{H\times W\times C}$.
In parallel, the image-wise skip module~$S$ takes $I_{LR}$ as input to generate a skipped feature~$F_{s}$, and the upsampler~$U$ utilize both $F$ and $F_{s}$ as inputs to return the upsampled image $I_{SR}\in\mathbb{R}^{rH\times rW\times 3}$ where $r$ denotes upscaling factor.
The entire process is formulated as follows:
\begin{equation}
\begin{aligned}
   F_{0} &= \mathrm{Conv}_{3\times3}(I_{LR}), \\
   F &= H(F_{0}, LK) + F_{0}, \\ 
   F_{s} &= S(I_{LR}), \\
   I_{SR} &= U(F, F_{s}).
\end{aligned}
\end{equation}
The $H$ is composed of the $N$ $\mathrm{ESCBlock}$s where $N$ denotes the number of $\mathrm{ESCBlock}$s and a 3$\times$3 convolution, and is formulated as follows:
\begin{equation}
\begin{aligned}
   F_{i} &= \mathrm{ESCBlock}_{i}(F_{i-1}, LK), \quad i=1,\dots, N,  \\ 
   F_{d} &= \mathrm{Conv}_{3\times3}(F_{N}),
\end{aligned}
\end{equation}
\noindent where $F_{i-1}$ indicates the feature input to the $i$th $\mathrm{ESCBlock}$ together with $LK$, and $F_{i}$ represents the extracted feature.
The $U$ and $S$ will be described in the implementation detail section because they will be changed depending on the task.

\subsection{ESC Block}
\begin{figure}[t]
  \centering
  \includegraphics[width=\columnwidth]{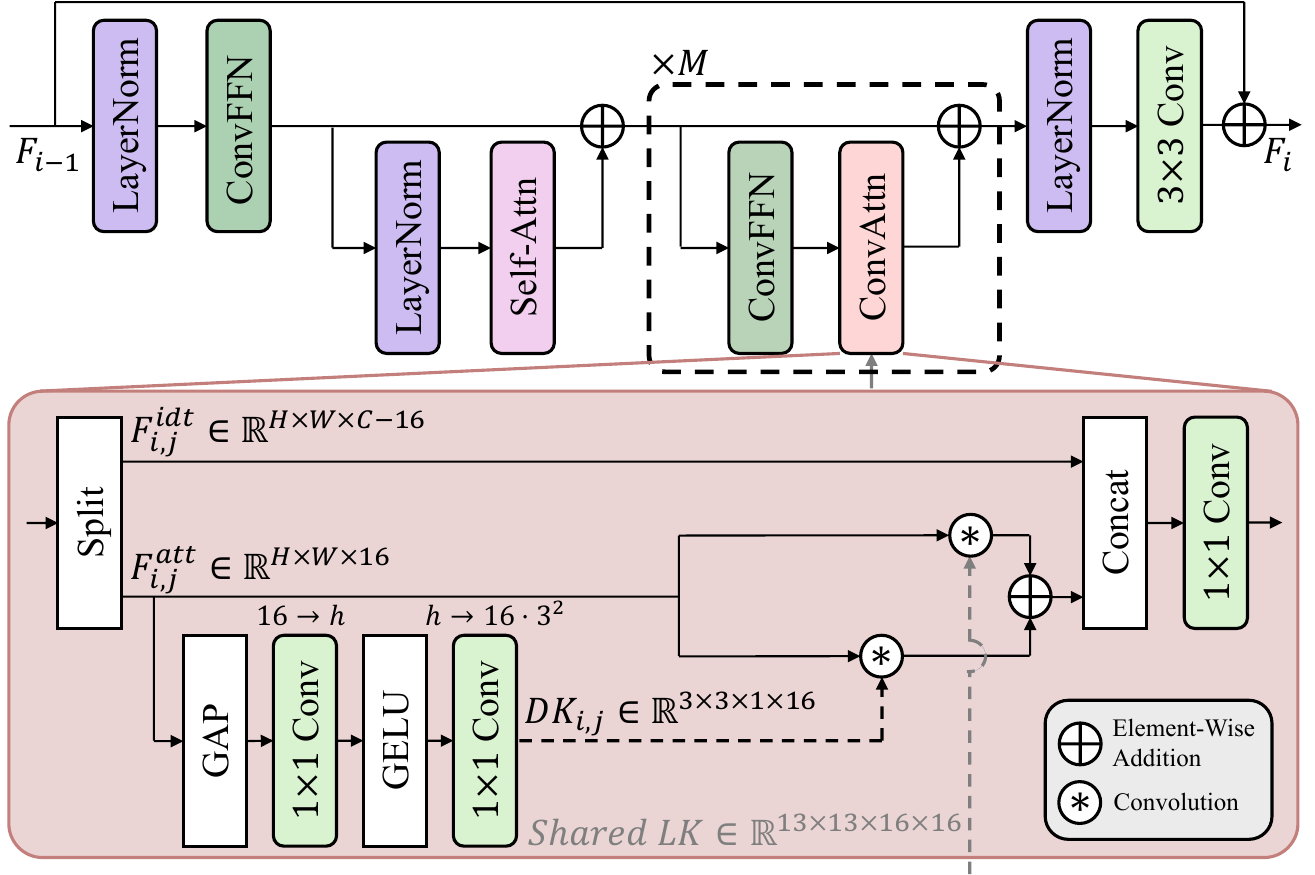}
  \vspace{-0.6cm}
  \caption{
    Visualization of the proposed ESC block.
  }
  \vspace{-0.6cm}
  \label{fig:block_module}
\end{figure}

The overall process of the $\mathrm{ESCBlock}$ is illustrated in Figure~\ref{fig:block_module} and is formulated as follows:
\begin{equation}
\begin{aligned}
   F^{in}_{i} &= \mathrm{ConvFFN}\bigl(\mathrm{LN}(F_{i-1})\bigr),\\
   F_{i, 0} &= F^{in}_{i} + \mathrm{SelfAttn}\bigl(\mathrm{LN}(F^{in}_{i})\bigr),\\
   F_{i, j} &= F_{i, j-1} + \mathrm{ConvAttn}_{j}\bigl(\mathrm{ConvFFN}_{j}(F_{i, j-1}), LK\bigr),\\
             &\quad \quad j=1,\dots,M,\\
   F_{i} &= F_{i-1} +\mathrm{Conv}_{3\times3}\bigl(\mathrm{LN}(F_{i, M})\bigr).
\end{aligned}
\end{equation}
Firstly, the $F_{i-1}$ is converted to the $F^{in}_{i}$ by layer normalization~($\mathrm{LN}$) and convolutional FFN~($\mathrm{ConvFFN}$)~\cite{SRFormer} with the kernel size of three. 
Then, $\mathrm{LN}$ and naive window self-attention~\cite{SwinIR} with the window size of 32$\times$32 are employed in a residual connection to extract $F_{i, 0}$.
Despite its large window size, we effectively reduce memory overhead by utilizing Flash Attention.
Furthermore, since $\mathrm{ConvFFN}$ is placed before, the self-attention extracts features considering local information without the complicated $QKV$ projection module. 
After establishing $F_{i, 0}$, the block iterates residual paths composed of $\mathrm{ConvFFN}$ and proposed ConvAttn module~($\mathrm{ConvAttn}$) $M$ times.
Although our methods do not employ shifted-window self-attention, these convolution-based modules after the self-attention will extract inter-window features. 
Finally, the result of the last iteration $F_{i, M}$ go through $\mathrm{LN}$ and $\mathrm{Conv}_{3\times3}$, then is added to $F_{i-1}$, resulting in $F_{i}$.

\noindent\textbf{Convolutional Attention Module}
The $\mathrm{ConvAttn}$ is designed to emulate self-attention's two key advantages: capturing long-range dependencies and instance-dependent weighting.
For this purpose, $\mathrm{ConvAttn}$ leverages two types of convolution formulated as:
\begin{equation}\label{eqn:CA}
\begin{aligned}
   F^{att}_{i, j}, F^{idt}_{i, j} &= \mathrm{Split}_{16:C-16}(F^{\mathrm{CF}}_{i, j}), \\
   DK_{i, j} &= \mathrm{Conv}^{up}_{1\times1}\bigl(\phi(\mathrm{Conv}^{down}_{1\times1}(\mathrm{GAP}(F^{att}_{i, j})))\bigr), \\
   F^{res}_{i, j} &= (F^{att}_{i, j} \circledast DK_{i,j}) + (F^{att}_{i, j} \circledast LK), \\
   F^{fuse}_{i, j} &= \mathrm{Conv}^{fuse}_{1\times1}\bigl(\mathrm{Concat}(F^{res}_{i, j}, F^{idt}_{i, j})\bigr),
\end{aligned}
\end{equation}
\noindent where $F^{\mathrm{CF}}_{i, j}$ denotes the features extracted by $\mathrm{ConvFFN}_{i,j}$ before.
We first split the $F^{\mathrm{CF}}_{i, j}$ channel-wise, resulting in $F^{att}_{i, j}\in\mathbb{R}^{H\times W\times 16}$, which consists of the 16 channels in front, and $F^{idt}_{i, j}\in\mathbb{R}^{H\times W\times C-16}$, which contains the remaining part.
Two types of convolutions are only applied within $F^{att}_{i, j}$ to reduce memory access cost, unlike the methods that leverage large kernels considering only computational complexity~\cite{ShuffleMixer, LKDN, ConvFormer}.
Next, dynamic depth-wise kernel $DK_{i, j}\in\mathbb{R}^{3\times3\times1\times16}$~\cite{DWNet} is created by two layers of $\mathrm{Conv}_{1\times1}$ with global average pooling~($\mathrm{GAP}$) and gaussian error linear unit activation~($\phi$).
Since enlarging $DK$'s kernel size increases the parameter size multiplied by the hidden dimension~($h$) of the kernel estimator, we fix its kernel size to 3$\times$3.
Then both $DK_{i, j}$ and $LK$ are convolved on $F^{att}_{i, j}$ and sum up those, resulting convolved feature $F^{res}_{i,j}$.
Based on our preliminary analysis, we hypothesize that long-range interactions do not significantly vary across layers.
Therefore, we share $LK$ across the entire $\mathrm{ESCBlock}$s and $\mathrm{ConvAttn}$s, mitigating parameter growth and optimization difficulty~\cite{SLaK, PeLK}.
Finally, $F^{res}_{i,j}$ and $F^{idt}_{i,j}$ are concatenated channel-wise, followed by the last convolution $\mathrm{Conv}^{fuse}_{1\times1}$ to fuse them, resulting in fused feature $F^{fuse}_{i, j}$.
In this way, both the local features extracted by $\mathrm{ConvFFN}$ and the long-range features extracted by $\mathrm{ConvAttn}$ can be leveraged to capture multi-scale information.

While $DK_{i, j}$ can be merged into $LK$ by padding both spatial and channel dimensions to reduce FLOPs, we do not because it slows down latency.
On the other hand, in situations where reducing FLOPs and parameter size is crucial rather than reducing latency, $LK$ can be decomposed depth-wise separable manner~\cite{MobileNetV1} into point-wise large kernel filter $LK^{\mathrm{c}}\in\mathbb{R}^{1\times1\times16\times16}$ and depth-wise large kernel filter $LK^{\mathrm{s}}\in\mathbb{R}^{13\times13\times1\times16}$.
This decomposition redefines line 3 in Equation~\ref{eqn:CA} as follows:
\begin{equation}
    F^{res}_{i, j} = (F^{att}_{i, j} \circledast LK^{c}) \circledast \bigr(ZP(DK_{i, j}) + LK^{s}\bigl),
\end{equation}
\noindent where $ZP(\cdot):\mathbb{R}^{3\times3\times1\times16}\mapsto\mathbb{R}^{13\times13\times1\times16}$ denotes spatial zero padding.

In summary, $\mathrm{ConvAttn}$ efficiently emulates self-attention's two advantages by leveraging a shared $LK$ and layer-specific $DK$s.
In contrast to Transformers that share attention maps across several layers~\cite{ELAN, ASID, UPS}, which still require a full materialized attention map for every layer, our module only leverages a single shared $LK$, significantly reducing memory footprints.
Moreover, unlike recent CNNs~\cite{ShuffleMixer, VapSR, LKDN, ConvFormer} that use large kernel convolutions and optionally employ pixel-wise attention to inject input dependency, we employ layer-specific $DK$s, emulating self-attention's input-dependent weight more exactly in a memory-efficient manner.

\begin{table}[!h]
\caption{
    Comparison of latency and memory usage between naive self-attention and our Flash Attention~(FA)~\cite{FlashAttn, FlexAttn}.
    WS and \#Pixels denote the window size of self-attention and the corresponding number of pixels.
    $X$ refers to the input tensor.
}\label{tab:attn_compare}
\vspace{-0.2cm}
\renewcommand{\arraystretch}{0.95}
\resizebox{\columnwidth}{!}{%
\begin{tabular}{@{}llcccc@{}}
\toprule
\multirow{3}{*}{\begin{tabular}[c]{@{}c@{}}WS\\(\#Pixels)\end{tabular}} & \multirow{3}{*}{Type} & \multicolumn{2}{c}{$X\in\mathbb{R}^{64\times64\times64}$} & \multicolumn{2}{c}{$X\in\mathbb{R}^{288\times288\times64}$} \\ \cmidrule(l){3-6} 
& & \begin{tabular}[c]{@{}c@{}}Latency\\ (ms)\end{tabular} & \begin{tabular}[c]{@{}c@{}}Memory Usage\\ (mb)\end{tabular} & \begin{tabular}[c]{@{}c@{}}Latency\\ (ms)\end{tabular} & \begin{tabular}[c]{@{}c@{}}Memory Usage\\ (mb)\end{tabular} \\ \midrule
\multirow{2}{*}{\begin{tabular}[c]{@{}c@{}}16$\times$16\\(256)\end{tabular}} & w/o FA & 2.0 & 46 & 4.9 & 758 \\
 & w/ FA & 0.5~\small{($4\times$)} & 11~\small{($4.2\times$)} & 1.3~\small{($3.8\times$)} & 223~\small{($3.4\times$)} \\ \midrule
 \multirow{2}{*}{\begin{tabular}[c]{@{}c@{}}24$\times$24\\(576)\end{tabular}} & w/o FA & 7.6 & 111 & 14.5 & 1575 \\
 & w/ FA & 0.5~\small{($15.2\times$)} & 14~\small{($7.9\times$)} & 2.0~\small{($7.3\times$)} & 223~\small{($7.1\times$)} \\ \midrule
\multirow{2}{*}{\begin{tabular}[c]{@{}c@{}}32$\times$32\\(1024)\end{tabular}} & w/o FA & 27.3 & 157 & 44.7 & 2717 \\
 & w/ FA & 0.5~\small{($54.6\times$)} & 11~\small{($14.3\times$)} & \textbf{2.8}~\small{\textcolor{myred}{($16.0\times$)}} & \textbf{223}~\small{\textcolor{myred}{($12.2\times$)}} \\ \bottomrule
\end{tabular}%
}
\vspace{-0.3cm}
\end{table}

\vspace{-0.15cm}

\begin{table*}[!ht]
\caption{
    Comparisons of classic SR methods trained on the DIV2K dataset. 
    Mem, \#FLOPs, and \#Params denote memory usage, the number of FLOPs, and parameter size, respectively.
    $\dagger$ denotes that we re-calculate the statistics of the indicated methods.
    All statistics and metrics are measured as detailed in Section~\ref{sec:classicsr_results}.
    The best result on each PSNR and SSIM is bolded.
}\label{tab:main_div2k}
\vspace{-0.3cm}
\renewcommand{\arraystretch}{0.95}
\resizebox{\textwidth}{!}{%
\begin{tabular}{@{}l|c|cccc|ccccc@{}}
\toprule
\multirow{2}{*}{Method} & \multirow{2}{*}{Scale} & \multirow{2}{*}{\begin{tabular}[c]{@{}c@{}}Latency\\ (ms)\end{tabular}} & \multirow{2}{*}{\begin{tabular}[c]{@{}c@{}}Mem\\ (mb)\end{tabular}} & \multirow{2}{*}{\begin{tabular}[c]{@{}c@{}}\#FLOPs\\ (G)\end{tabular}} & \multirow{2}{*}{\begin{tabular}[c]{@{}c@{}}\#Params\\ (K)\end{tabular}} & \multicolumn{5}{c}{PSNR / SSIM} \\
 &  &  &  &  &  & Set5 & Set14 & B100 & Urban100 & Manga109 \\ \midrule
SwinIR-lt~\cite{SwinIR}& \multirow{13}{*}{$\times2$} & 1409.8 & 1287 & 244.2 & 910 & 38.14/0.9611 & 33.86/0.9206 & 32.31/0.9012 & 32.76/0.9340 & 39.12/0.9783 \\
ELAN-lt~\cite{ELAN} &  & 94.5 & 887 & 203.1 & 621 & 38.17/0.9611 & 33.94/0.9207 & 32.30/0.9012 & 32.76/0.9340 & 39.11/0.9782 \\
OmniSR~\cite{OmniSR} &  & 120.3 & 1031 & 194.5 & 772 & 38.22/0.9613 & 33.98/0.9210 & 32.36/0.9020 & 33.05/0.9363 & 39.28/0.9784 \\
SRFormer-lt~\cite{SRFormer} &  & 1456.3 & 1184 & 236.3 & 853 & 38.23/0.9613 & 33.94/0.9209 & 32.36/0.9019 & 32.91/0.9353 & 39.28/0.9785 \\
ATD-lt~\cite{ATD} &  & 733.5 & 2839 & 380.0 & 753 & 38.29/0.9616 & 34.10/0.9217 & 32.39/0.9023 & 33.27/0.9375 & 39.52/0.9789 \\
HiT-SRF~\cite{HiTSR} &  & 268.1 & 1804 & 226.5 & 847 & 38.26/0.9615 & 34.01/0.9214 & 32.37/0.9023 & 33.13/0.9372 & 39.47/0.9787 \\
ASID-D8~\cite{ASID} &  & 131.2 & 999 & 190.5$^{\dagger}$ & 732 & 38.32/0.9618 & \textbf{34.24}/\textbf{0.9232} & 32.40/\textbf{0.9028} & 33.35/0.9387 & -~/~- \\
MambaIR-lt~\cite{MambaIR} &  & 277.1 & 1695 & 334.2 & 905 & 38.13/0.9610 & 33.95/0.9208 & 32.31/0.9013 & 32.85/0.9349 & 39.20/0.9782 \\
MambaIRV2-lt~\cite{MambaIRV2} &  & 580.4 & 2824 & 286.3 & 774 & 38.26/0.9615 & 34.09/0.9221 & 32.36/0.9019 & 33.26/0.9378 & 39.35/0.9785 \\
RDN~\cite{RDN} &  & 279.3 & 2058 & 5096.2 & 22123 & 38.24/0.9614 & 34.01/0.9212 & 32.34/0.9017 & 32.89/0.9353 & 39.18/0.9780 \\
RCAN~\cite{RCAN} &  & 299.3 & 626 & 3529.7 & 15445 & 38.27/0.9614 & 34.12/0.9216 & \textbf{32.41}/0.9027 & 33.34/0.9384 & 39.44/0.9786 \\
\textbf{ESC-FP~(Ours)} &  & 94.3 & 627 & 239.8 & 524 & 38.27/0.9617 & 34.09/0.9219 & 32.37/0.9022 & 33.22/0.9375 & 39.40/0.9784 \\
\textbf{ESC-lt~(Ours)} &  & 73.2 & 830 & 359.4 & 603 & 38.24/0.9615 & 33.98/0.9211	& 32.35/0.9020 & 33.05/0.9363 & 39.33/0.9786 \\
\textbf{ESC~(Ours)} &  & 120.9 & 831 & 592.0 & 947 & \textbf{38.35}/\textbf{0.9619} & 34.11/0.9223 & \textbf{32.41}/0.9027 & \textbf{33.46}/\textbf{0.9395} & \textbf{39.54}/\textbf{0.9790} \\ \midrule
SwinIR-lt~\cite{SwinIR}& \multirow{13}{*}{$\times3$} & 331.7 & 596 & 110.8 & 918 & 34.62/0.9289 & 30.54/0.8463 & 29.20/0.8082 & 28.66/0.8624 & 33.98/0.9478 \\
ELAN-lt~\cite{ELAN} &  & 32.5 & 399 & 90.1 & 629 & 34.61/0.9288 & 30.55/0.8463 & 29.21/0.8081 & 28.69/0.8624 & 34.00/0.9478 \\
OmniSR~\cite{OmniSR} &  & 41.2 & 476 & 88.4 & 780 & 34.70/0.9294 & 30.57/0.8469 & 29.28/0.8094 & 28.84/0.8656 & 34.22/0.9487 \\
SRFormer-lt~\cite{SRFormer} &  & 530.5 & 537 & 105.4 & 861 & 34.67/0.9296 & 30.57/0.8469 & 29.26/0.8099 & 28.81/0.8655 & 34.19/0.9489 \\
ATD-lt~\cite{ATD} &  & 274.4 & 1258 & 168.0 & 760 & 34.74/0.9300 & 30.68/0.8485 & 29.32/0.8109 & 29.17/0.8709 & 34.60/0.9506 \\
HiT-SRF~\cite{HiTSR} &  & 124.9 & 1464 & 101.6 & 855 & 34.75/0.9300 & 30.61/0.8475 & 29.29/0.8106 & 28.99/0.8687 & 34.53/0.9502 \\
ASID-D8~\cite{ASID} &  & 61.9 & 460 & 86.4$^{\dagger}$ & 739 & \textbf{34.84}/0.9307 & 30.66/0.8491 & 29.32/\textbf{0.8119} & 29.08/0.8706 & -~/~- \\
MambaIR-lt~\cite{MambaIR} &  & 109.3 & 760 & 148.5 & 913 & 34.63/0.9288 & 30.54/0.8459 & 29.23/0.8084 & 28.70/0.8631 & 34.12/0.9479 \\
MambaIRV2-lt~\cite{MambaIRV2} &  & 259.0 & 1250 & 126.7 & 781 & 34.71/0.9298 & 30.68/0.8483 & 29.26/0.8098 & 29.01/0.8689 & 34.41/0.9497 \\
RDN~\cite{RDN} &  & 146.5 & 985 & 2281.2 & 22308 & 34.71/0.9296 & 30.57/0.8468 & 29.26/0.8093 & 28.80/0.8653  & 34.13/0.9484 \\
RCAN~\cite{RCAN} &  & 85.1 & 560 & 1586.1 & 15629 & 34.74/0.9299 & 30.65/0.8482 & 29.32/0.8111 & 29.09/0.8702 & 34.44/0.9499 \\
\textbf{ESC-FP~(Ours)} &  & 34.8 & 291 & 110.0 & 530 & 34.72/0.9300 & 30.67/0.8483 & 29.30/0.8107 &  29.12/0.8706 & 34.56/0.8706 \\
\textbf{ESC-lt~(Ours)} &  & 25.2 & 384 & 162.8 & 612 & 34.61/0.9295 & 30.52/0.8475 & 29.26/0.8102 & 28.93/0.8679 & 34.33/0.9495 \\
\textbf{ESC~(Ours)} &  & 41.4 & 385 & 267.6 & 955 & \textbf{34.84}/\textbf{0.9308} & \textbf{30.74}/\textbf{0.8493} & \textbf{29.34}/0.8118 & \textbf{29.28}/\textbf{0.8739} & \textbf{34.66}/\textbf{0.9512} \\ \midrule
SwinIR-lt~\cite{SwinIR}& \multirow{13}{*}{$\times4$} & 222.9 & 351 & 63.6 & 930 & 32.44/0.8976 & 28.77/0.7858 & 27.69/0.7406 & 26.47/0.7980 & 30.92/0.9151 \\
ELAN-lt~\cite{ELAN} &  & 18.0 & 241 & 54.1 & 640 & 32.43/0.8975 & 28.78/0.7858 & 27.69/0.7406 & 26.54/0.7982 & 30.92/0.9150 \\
OmniSR~\cite{OmniSR} &  & 22.5 & 273 & 50.9 & 792 & 32.49/0.8988 & 28.78/0.7859 & 27.71/0.7415 & 26.64/0.8018 & 31.02/0.9151 \\
SRFormer-lt~\cite{SRFormer} &  & 287.2 & 329 & 62.8 & 873 & 32.51/0.8988 & 28.82/0.7872 & 27.73/0.7422 & 26.67/0.8032 & 31.17/0.9165 \\
ATD-lt~\cite{ATD} &  & 189.7 & 753 & 100.1 & 769 & 32.63/0.8998 & 28.89/0.7886 & 27.79/0.7440 & 26.97/0.8107 & 31.48/0.9198 \\
HiT-SRF~\cite{HiTSR} &  & 82.1 & 1331 & 58.0 & 866 & 32.55/0.8999 & 28.87/0.7880 & 27.75/0.7432 & 26.80/0.8069 & 31.26/0.9171 \\
ASID-D8~\cite{ASID} &  & 61.8 & 265 & 49.6$^{\dagger}$ & 748 & 32.57/0.8990 & 28.89/0.7898 & 27.78/0.7449 & 26.89/0.8096 & -~/~- \\
MambaIR-lt~\cite{MambaIR} &  & 55.8 & 438 & 84.6 & 924 & 32.42/0.8977 & 28.74/0.7847 & 27.68/0.7400 & 26.52/0.7983 & 30.94/0.9135\\
MambaIRV2-lt~\cite{MambaIRV2} &  & 153.4 & 748 & 75.6 & 790 & 32.51/0.8992 & 28.84/0.7878 & 27.75/0.7426 & 26.82/0.8079 & 31.24/0.9182 \\
RDN~\cite{RDN} &  & 66.0 & 791 & 1309.2 & 22271 & 32.47/0.8990 & 28.81/0.7871 & 27.72/0.7419 & 26.61/0.8028 & 31.00/0.9151 \\
RCAN~\cite{RDN} &  & 52.2 & 540 & 917.6 & 15592 & 32.63/0.9002 & 28.87/0.7889 & 27.77/0.7436 & 26.82/0.8087 & 31.22/0.9173 \\
\textbf{ESC-FP~(Ours)} &  & 21.7 & 158 & 60.8 & 539 & 32.56/0.8897 & 28.87/0.7889 & 27.75/0.7435 & 26.90/0.8098 & 31.40/0.9192 \\
\textbf{ESC-lt~(Ours)} &  & 13.3 & 213 & 91.0 & 624 & 32.52/0.8995 & 28.87/0.7878 & 27.72/0.7423 & 26.76/0.8058 & 31.26/0.9173 \\
\textbf{ESC~(Ours)} &  & 21.9 & 215 & 149.2 & 968 & \textbf{32.68}/\textbf{0.9011} & \textbf{28.93}/\textbf{0.7902} & \textbf{27.80}/\textbf{0.7447} & \textbf{27.07}/\textbf{0.8144} & \textbf{31.54}/\textbf{0.9207} \\ \bottomrule
\end{tabular}%
}
\vspace{-0.5cm}
\end{table*}

\section{Experiments}
This section demonstrates the effectiveness of our flash attention and our extensive experiments on various tasks.
For implementation details for proposed networks, please refer to Section~\ref{sec:implementation_details} in the supplementary.

\subsection{Effectiveness of Flash Attention}
We address the excessive memory overhead of self-attention by leveraging Flash Attention~\cite{FlashAttn, FlexAttn}.
Implementation details for Flash Attention are described in Section~\ref{sec:fa_imp_det} in the supplementary.
As shown in Table~\ref{tab:attn_compare}, our self-attention implementation accelerated by Flash Attention significantly reduces the latency and memory usage of self-attention at the window size of 32 by 16$\times$ and 12.2$\times$, respectively.
We do not consider window sizes above 32, as it may cause training instability due to the necessity of padding, since the training patch size is 64$\times$64.

\begin{figure}[ht]
  \centering
  \includegraphics[width=\columnwidth]{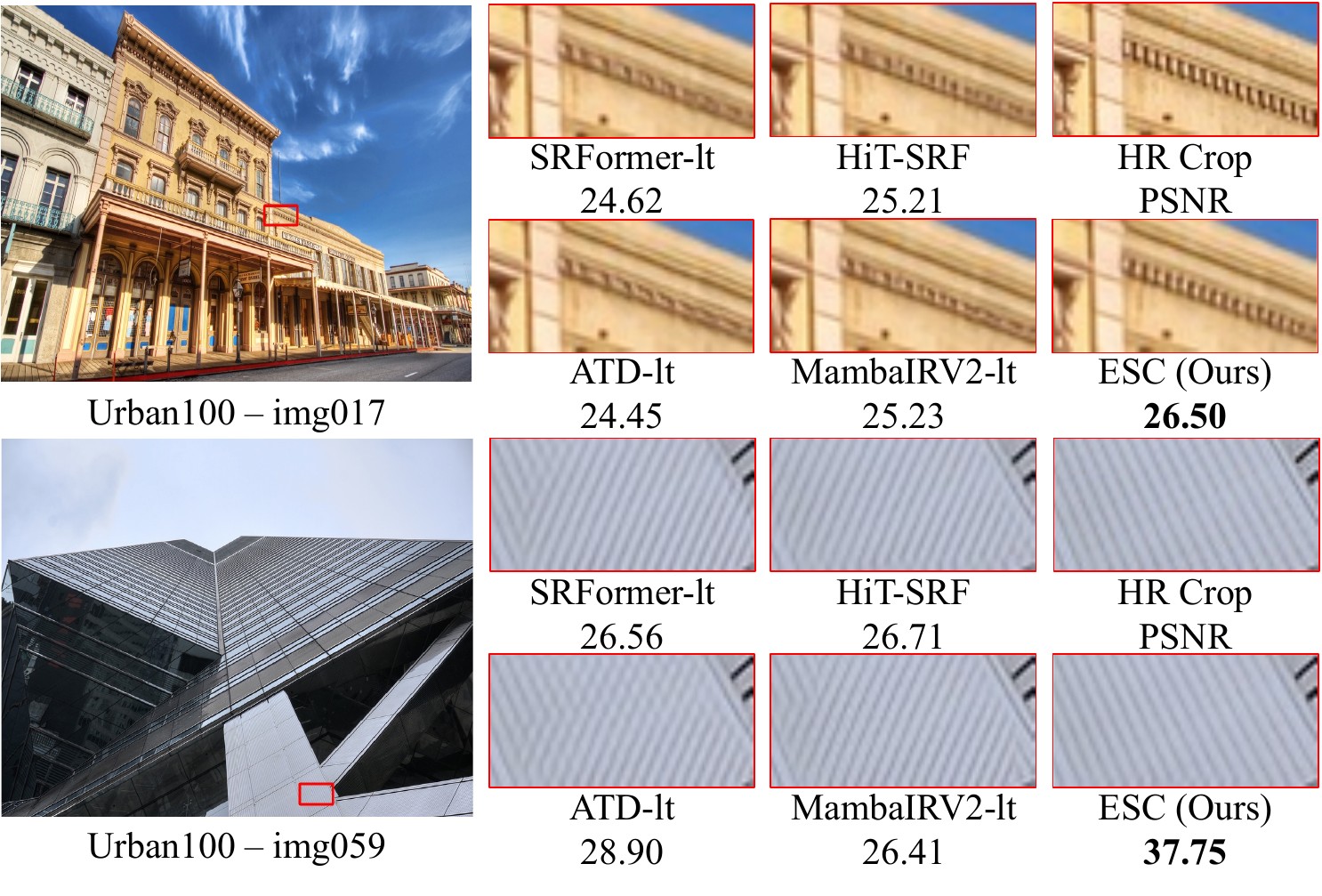}
  \vspace{-0.8cm}
  \caption{
    Visual comparisons for classic SR~($\times$2).
    The best result on PSNR is bolded.
  }
  \label{fig:classic_visual}
  \vspace{-0.6cm}
\end{figure}

\subsection{Classic SR Results}\label{sec:classicsr_results}
To evaluate our proposed networks on the classic SR task, we leverage five commonly used datasets~(Set5~\cite{Set5}, Set14~\cite{Set14}, B100~\cite{BSD100}, Urban100~\cite{Urban100}, Manga109~\cite{Manga109}). 
We measure peak signal-to-noise ratio~(PSNR) and structural similarity index measure~(SSIM) on the Y channel after cropping the boundary corresponding to the upscaling factor and converting them into YCbCr space.
All statistics are measured by reconstructing an HD image~(1280$\times$720) image using an RTX4090 GPU at FP32 precision 100 times.
Memory usage is calculated by \texttt{torch.cuda.max\_memory\_allocated} function and FLOPs are measured by \texttt{fvcore} library.
For the classic SR task, we present three variants: ESC, ESC-light, and ESC-FP.
ESC and ESC-light prioritize reducing latency, while ESC-FP focuses on reducing FLOPs and parameter size.

\subsubsection{Quantitative Results on the DIV2K dataset}
We compare our methods with various representative lightweight SR methods, including CNN-based models~\cite{RDN, RCAN}, Transformers~\cite{SwinIR, ELAN, OmniSR, SRFormer, ATD, HiTSR, ASID}, and SSMs~\cite{MambaIR, MambaIRV2} trained on the DIV2K dataset~\cite{DIV2K}.
Notably, as shown in Table~\ref{tab:main_div2k}, our ESC achieves a notable 0.1dB PSNR improvement on the Urban100$\times$4 dataset, while being 8.7$\times$ faster than ATD-light.
Moreover, ESC-light is 1.4$\times$ faster than ELAN-light, exhibiting an impressive performance improvement of 0.34dB PSNR in the Manga109$\times4$ dataset.
These results exhibit that state-of-the-art performance can be achieved even when most self-attentions are replaced with convolutions, suggesting that replacing self-attention with convolutions does not compromise representational power.
Furthermore, ESC-FP outperforms MambaIRV2-light over 0.18dB on the Urban100$\times4$ dataset while reducing FLOPs and parameters by 20\% and 31\%, respectively. 
This result underscores that our design is effective even when reducing FLOPs and parameter size.

\subsubsection{Visual Results on the DIV2K dataset}
We also visually compare our ESC with representative lightweight SR methods, including SRFormer-light~\cite{SRFormer}, HiT-SRF~\cite{HiTSR}, ATD-light~\cite{ATD}, and MambaIRV2-light~\cite{MambaIRV2}.
As shown in Figure~\ref{fig:classic_visual}, our ESC effectively restores details and achieves a high PSNR, demonstrating that our network also produces visually pleasant results.

\begin{table}[!ht]
\caption{
    Ablation studies on our ESC network. 
    All comparisons are designed to have similar latencies to ESC. 
    $LK$, $DK$, and WS denote large kernel, dynamic kernel, and window size, respectively.
    All statistics and metrics are measured as detailed in Section~\ref{sec:classicsr_results}.
}\label{tab:ablation}
\vspace{-0.2cm}
\resizebox{\columnwidth}{!}{
\begin{tabular}{@{}l|cccc|cc@{}}
\toprule
Cases & \begin{tabular}[c]{@{}c@{}}Latency\\ (ms)\end{tabular} & \begin{tabular}[c]{@{}c@{}}Memory\\ (mb)\end{tabular} & \begin{tabular}[c]{@{}c@{}}\#FLOPs\\ (G)\end{tabular} & \begin{tabular}[c]{@{}c@{}}\#Params\\ (K)\end{tabular} & Set5$\times$2 & Urban100$\times$2 \\ \midrule
Only Self-attentions & 128.2 & 830 & 456.6 & 737 & 38.27/0.9617 & 33.23/0.9378 \\
Only ConvAttns & 126.3 & 514 & 962.4 & 1210 & 38.18/0.9613 & 32.91/0.9353  \\ \midrule
9$\times$9 $LK$ & 117.1 & 831 & 462.2 & 924 & 38.33/0.9618 & 33.42/0.9392 \\
17$\times$17 $LK$ & 126.6 & 832 & 768.9 & 977 & 38.32/0.9618 & 33.40/0.9390 \\ \midrule
w/o $LK$ Share and $DK$ & 119.7 & 835 & 591.1 & 1950 & 38.32/0.9618 & 33.37/0.9389 \\
w/o $LK$ Share & 120.9 & 835 & 592.0 & 1985 & 38.32/0.9619 & 33.43/0.9392 \\
w/o $DK$ & 119.7 & 835 & 591.1 & 911 & 38.31/0.9618 & 33.36/0.9388 \\\midrule
WS16 and more layers & 118.0 & 807 & 603.6 & 1054 & 38.24/0.9614 & 33.05/0.9360 \\
WS24 and more layers & 116.5 & 801 & 583.5 & 995 & 38.33/0.9618 & 33.28/0.9383 \\ \midrule
\textbf{ESC~(Ours)} & 120.9 & 831 & 592.0 & 947 & 38.35/0.9619 & 33.46/0.9395  \\ \bottomrule
\end{tabular}%
}
\vspace{-0.5cm}
\end{table}

\subsubsection{Ablation Study}
We conduct an ablation study to justify our network's composition, as shown in Table~\ref{tab:ablation}.
First, we introduce two variants, one that uses only the self-attention module and another that uses only the $\mathrm{ConvAttn}$.
The number of blocks and layers is adjusted to match the latency on par with the ESC.
Both variants exhibit a performance drop compared to ESC, suggesting that leveraging both self-attention and $\mathrm{ConvAttn}$ offers the best performance-efficiency trade-off.
Next, we vary the $LK$'s kernel size to 9$\times$9 and 17$\times$17 to determine the optimal size. 
In both cases, we observe a drop in performance, confirming that a kernel size of 13$\times$13 is optimal.
Then, we evaluate performance changes by altering the convolution type used in the $\mathrm{ConvAttn}$. 
First, we observe performance drops when neither $LK$ sharing nor $DK$ is used.
Then, we add $DK$ without sharing $LK$ and observe some performance improvement, though it still falls short of ESC.
This result suggests that addressing the optimization difficulty that comes from the massive parameter size is more critical than the layer-specific long-range modeling.
Finally, we leverage only $LK$ sharing without $DK$ and observe significant performance drops, affirming that harmonizing both $LK$ and $DK$ is crucial.
For the next step, we investigate the effect of the self-attention window size by setting it to 16 and 24 and adding extra layers. 
In every case, we observe a performance drop, which confirms that enlarging the window size to enhance self-attention is effective even when only a single self-attention is leveraged in each block.

\begin{table*}[!ht]
\caption{
    Comparisons of classic SR methods trained on the large-scale DFLIP dataset. 
    Mem, \#FLOPs, and \#Params denote memory usage, the number of FLOPs, and parameter size, respectively.
    $\S$ denotes that we train the indicated methods following their descriptions.
    All statistics and metrics are measured as detailed in Section~\ref{sec:classicsr_results}.
    The best result on each PSNR and SSIM is bolded.
}\label{tab:fixedscale_dflip}
\vspace{-0.3cm}
\resizebox{\textwidth}{!}{%
\begin{tabular}{@{}l|c|cccc|ccccc@{}}
\toprule
\multirow{2}{*}{Method} & \multirow{2}{*}{Scale} & \multirow{2}{*}{\begin{tabular}[c]{@{}c@{}}Latency\\ (ms)\end{tabular}} & \multirow{2}{*}{\begin{tabular}[c]{@{}c@{}}Mem\\ (mb)\end{tabular}} & \multirow{2}{*}{\begin{tabular}[c]{@{}c@{}}\#FLOPs\\ (G)\end{tabular}} & \multirow{2}{*}{\begin{tabular}[c]{@{}c@{}}\#Params\\ (K)\end{tabular}} & \multicolumn{5}{c}{PSNR / SSIM} \\
 &  &  &  &  &  & Set5 & Set14 & B100 & Urban100 & Manga109 \\ \midrule
SRFormer-lt$^{\S}$~\cite{SRFormer} & \multirow{4}{*}{$\times2$} & 1838.1 & 1184 & 236.3 & 853 & 38.24/0.9615 & 34.13/0.9218 & 32.42/0.9026 & 33.37/0.9386 & 39.36/0.9787 \\
ATD-lt$^{\S}$~\cite{ATD} &  & 733.5 & 2839 & 380.0 & 753 & 38.29/0.9616 & 34.30/0.9230 & 32.43/0.9027 & 33.62/0.9401 & 39.60/0.9791 \\
HiT-SRF$^{\S}$~\cite{HiTSR} &  & 268.1 & 1804 & 226.5 & 847 & 38.31/0.9616 & 34.31/0.9230 & 32.45/0.9031 & 33.58/0.9404 & 39.69/0.9793 \\
\textbf{ESC~(Ours)} &  & 120.9 & 831 & 592.0 & 947 & \textbf{38.34}/\textbf{0.9618} & \textbf{34.42}/\textbf{0.9235} & \textbf{32.50}/\textbf{0.9036} & \textbf{33.86}/\textbf{0.9424} & \textbf{39.73}/\textbf{0.9795} \\ \midrule
SRFormer-lt$^{\S}$~\cite{SRFormer} & \multirow{4}{*}{$\times3$} & 668.3 & 537 & 105.4 & 861 & 34.67/0.9297 & 30.75/0.8484 & 29.30/0.8108 & 29.10/0.8701 & 34.26/0.9498 \\
ATD-lt$^{\S}$~\cite{ATD} &  & 274.4 & 1258 & 168.0 & 760 & 34.71/0.9300 & 30.77/0.8493 & 29.33/0.8116 & 29.42/0.8743 & 34.61/0.9509 \\
HiT-SRF$^{\S}$~\cite{HiTSR} &  & 124.9 & 1464 & 101.6 & 855 & 34.69/0.9298 & 30.81/0.8493 & 29.32/0.8115 & 29.28/0.8729 & 34.72/0.9511 \\
\textbf{ESC~(Ours)} &  & 41.4 & 385 & 267.6 & 955 & \textbf{34.85}/\textbf{0.9312} & \textbf{30.97}/\textbf{0.8511} & \textbf{29.41}/\textbf{0.8135} & \textbf{29.70}/\textbf{0.8799} & \textbf{34.94}/\textbf{0.9525} \\ \midrule
SRFormer-lt$^{\S}$~\cite{SRFormer} & \multirow{4}{*}{$\times4$} & 327.8 & 329 & 62.8 & 873 & 32.49/0.8993 & 28.89/0.7887 & 27.76/0.7429 & 26.90/0.8086 & 31.25/0.9189 \\
ATD-lt$^{\S}$~\cite{ATD} &  & 189.7 & 753 & 100.1 & 769 & 32.52/0.8995 & 28.93/0.7896 & 27.79/0.7443 & 27.18/0.8150 & 31.47/0.9208 \\
HiT-SRF$^{\S}$~\cite{HiTSR} &  & 82.1 & 1331 & 58.0 & 866 & 32.55/0.8997 & 28.96/0.7897 & 27.77/0.7443 & 27.07/0.8130 & 31.59/0.9208 \\
\textbf{ESC~(Ours)} &  & 21.9 & 215 & 149.2 & 968 & \textbf{32.79}/\textbf{0.9025} & \textbf{29.06}/\textbf{0.7927} & \textbf{27.85}/\textbf{0.7466} & \textbf{27.45}/\textbf{0.8229} & \textbf{31.87}/\textbf{0.9239} \\ \bottomrule
\end{tabular}%
}
\vspace{-0.4cm}
\end{table*}

\subsubsection{Analysis on ConvAttn Module}
\begin{figure}[t]
  \centering
  \includegraphics[width=\columnwidth]{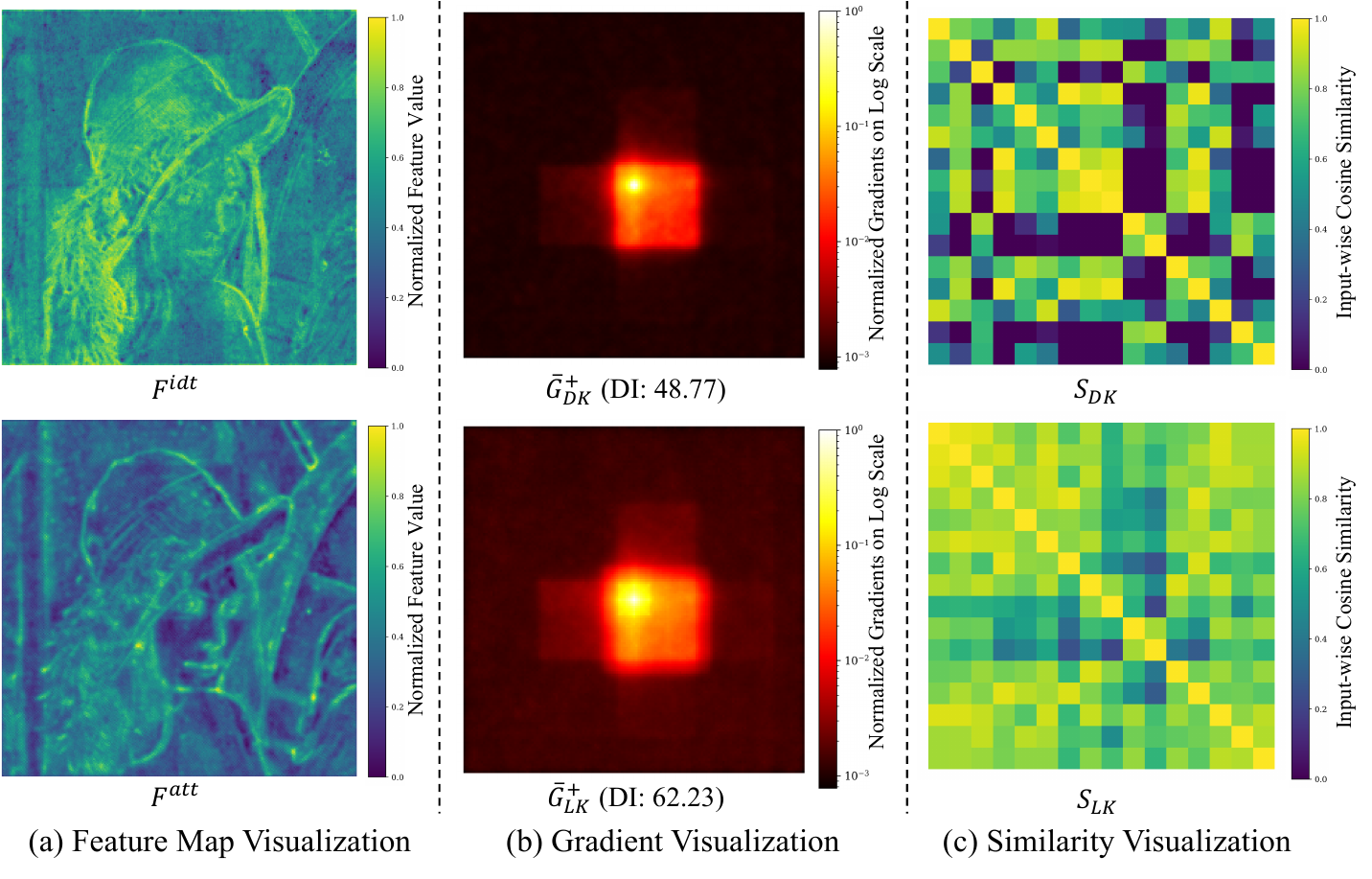}
  \vspace{-0.8cm}
  \caption{
    Visualized feature map, gradients, and input-wise similarity. $F$ is averaged across entire $\mathrm{ConvAttn}$ while processing the Lenna image in Set14. $\bar{G}^{+}$ and $S$ are calculated on the Urban100, and 10 samples in front of the Urban100, respectively. 
  }
  \label{fig:frequency}
  \vspace{-0.4cm}
\end{figure}

This section verifies that the proposed $\mathrm{ConvAttn}$ operates similarly to self-attention.
Self-attention is known to extract shape-biased structural features~\cite {HowDoViTWorks, RepLKNet}, unlike 3$\times$3 convolution, which mainly extracts local texture and edges.
Indeed, as shown in Figure~\ref{fig:frequency} (a), $F^{idt}$ focuses on local textures and edges while $F^{att}$ concentrates on structural components.
Next, to verify each convolution's receptive field, we visualize positive gradients $\bar{G}^{+}=\tfrac{1}{NMC}\sum_{p=1}^{NMC}\max\bigl(0,\tfrac{\partial I^{SR}_{i,j}}{\partial F_p}\bigr)$ propagated from the output image’s center pixel $(i,j)$ to the inputs feature~($F$) of both $DK$ and $LK$ across $N$ blocks, $M$ modules, and $C$ channels, and corresponding diffusion index~(DI).
Figure~\ref{fig:frequency} (b) clearly demonstrates that $LK$ extracts more long-range features than $DK$.
Lastly, we visualize input-wise similarity~($S$) between averaged pool features extracted by $LK$ and $DK$.
As demonstrated in Figure~\ref{fig:frequency} (c), $S_{DK}$ exhibits higher input-dependent diversity than $S_{LK}$, confirming that $DK$ extracts more input-dependent features than $LK$.

\begin{figure}[ht]
  \centering
  \includegraphics[width=\columnwidth]{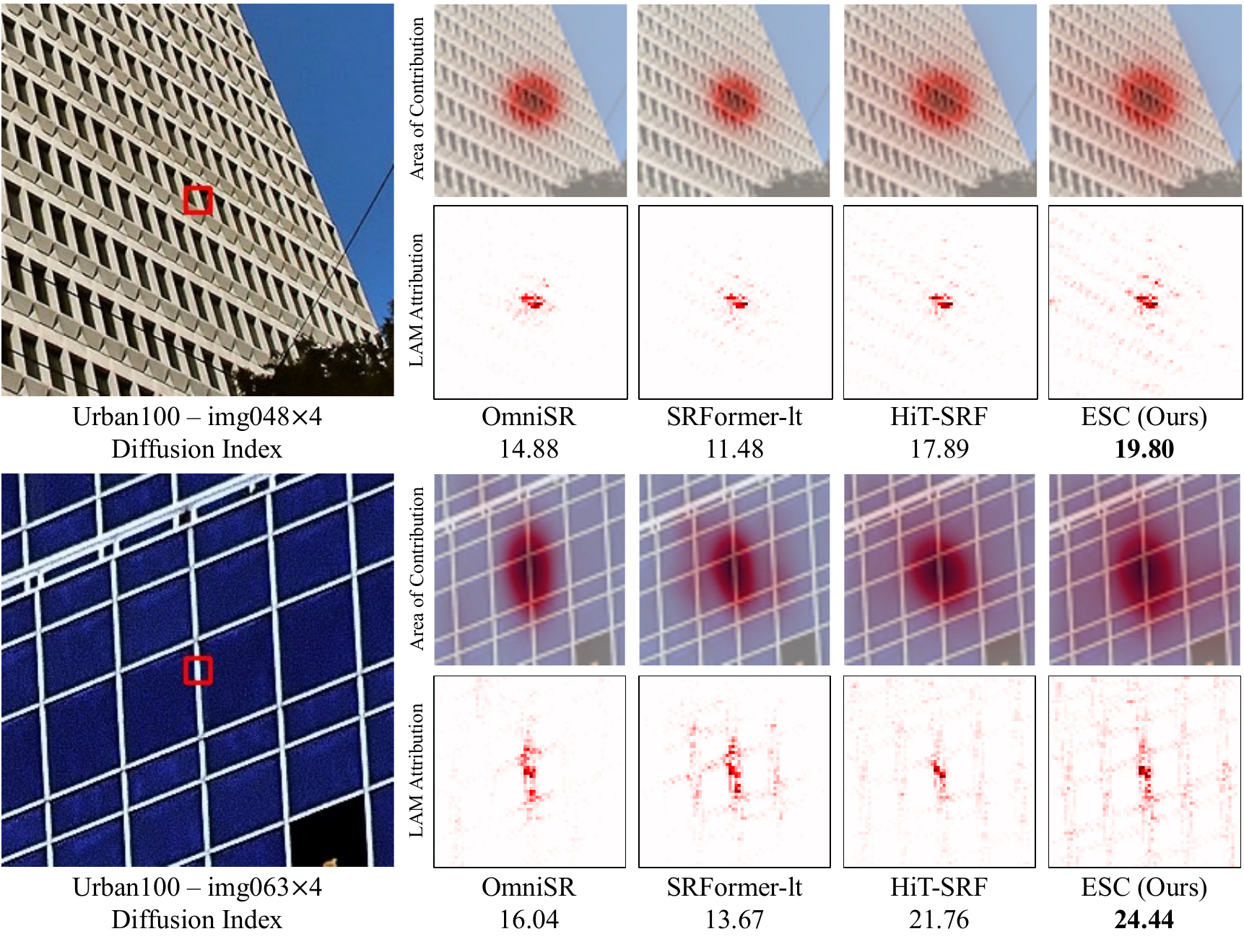}
  \vspace{-0.8cm}
  \caption{
    Comparisons of Local Attribution Map~(LAM)~\cite{LAM} and the corresponding area of contribution and diffusion index.
    The best result on the diffusion index is bolded.
  }
  \label{fig:lam}
  \vspace{-0.4cm}
\end{figure}

\subsubsection{Local Attribution Map}
We use the Local Attribution Map (LAM)~\cite{LAM} to verify that our networks can extract long-range dependencies despite using less self-attention.
We compare our ESC with studies of various self-attention implementations that aim to reduce FLOPs while expanding the receptive field, such as global-level attention~\cite{OmniSR}, permuted self-attention~\cite{SRFormer}, and spatial-channel correlation~\cite{HiTSR}.
Notably, as described in Figure~\ref{fig:lam}, our ESC references the widest range of pixels to restore the red bounding box along with the highest diffusion index.
This result exhibits that our network effectively captures long-range dependencies even though most self-attentions are replaced by the ConvAttn module, suggesting that our ConvAttn module contributes to retaining the large receptive field of Transformers.

\subsubsection{Quantitative Results on the DFLIP dataset}
Transformers exhibit significant performance improvements when trained on large-scale datasets~\cite{ViT, IPT}. 
To verify our model also retains this benefit, we train and evaluate various representative Transformers -- including SRFormer-light~\cite{SRFormer}, ATD-light~\cite{ATD}, and HiT-SRF~\cite{HiTSR} -- alongside our ESC on a large dataset.
For this purpose, we leverage DIV2K+Flickr2K+LSDIR+DiverSeg-IP~(DFLIP)~\cite{DIV2K, DF2KDataset, LSDIR, DiverSeg} dataset.
As illustrated in Table~\ref{tab:fixedscale_dflip}, our approach outperforms the other methods by a substantial margin across all scales and evaluation datasets. 
These results indicate that replacing self-attention with $\mathrm{ConvAttn}$ does not sacrifice the data scaling capability of Transformers.

\subsection{Arbitrary-Scale SR}
To evaluate our proposed networks on arbitrary-scale SR tasks, we leverage the DIV2K-validation~\cite{DIV2K} dataset.
Following previous research~\cite{LIIF, LTE}, we assess PSNR in the RGB channel after cropping the boundaries according to the upscaling factor plus six.
For arbitrary-scale SR tasks, we change $U$ of our network and representative lightweight SR transforemrs~\cite{ATD, HiTSR} into LTE~\cite{LTE}.

\subsubsection{Quantitative Results}
To demonstrate the outstanding representational capability of the proposed network, we train and evaluate the network on an arbitrary-scale SR task, which requires processing various upscaling factors with a single network.
For quantitative comparison, we leverage commonly used RDN~\cite{RDN} and additional representative Transformers~\cite{ATD, HiTSR} that we train.
As described in Tables~\ref{tab:fixedscale_arb_div}, our network continually achieves high performance at the scale both seen and unseen during training, even achieving a 0.09dB improvement over RDN+LTE at $\times$2. 
This result underscores our ESC's exceptional representational capability.
For quantitative results measured on other datasets, including Set5, Set14, B100, and Urban100, refer to Section~\ref{sec:add_res_assr} in the supplementary.

\begin{table}[!ht]
\caption{
    Quantitative comparison on arbitrary-scale SR task employing LTE~\cite{LTE} as upsampler. The best result on PSNR is bolded.
}\label{tab:fixedscale_arb_div}
\vspace{-0.3cm}
\resizebox{\columnwidth}{!}{%
\begin{tabular}{@{}l|ccc|ccccc@{}}
\toprule
\multirow{2}{*}{Methods} & \multicolumn{3}{c|}{Seen} & \multicolumn{5}{c}{Unseen} \\ \cmidrule(l){2-9} 
 & $\times$2 & $\times$3 & $\times$4 & $\times$6 & $\times$12 & $\times$18 & $\times$24 & $\times$30 \\ \midrule
RDN+LTE~\cite{RDN} & 35.04 & 31.32 & 29.33 & 27.04 & 23.95 & \textbf{22.40} & 21.36 & 20.64 \\
ATD-lt+LTE$^{\S}$~\cite{ATD} & 34.98 & 31.25 & 29.27 & 27.01 & 23.94 & 22.38 & 21.37 & 20.64 \\
HiT-SRF+LTE$^{\S}$~\cite{HiTSR} & 35.10 & 31.35 & 29.35 & 27.05 & 23.94 & 22.39 & 21.36 & 20.64 \\
\textbf{ESC+LTE~(Ours)} & \textbf{35.13} & \textbf{31.38} & \textbf{29.37} & \textbf{27.08} & \textbf{23.97} & \textbf{22.40} & \textbf{21.38} & 20.64 \\ \bottomrule
\end{tabular}%
}
\vspace{-0.3cm}
\end{table}

\subsubsection{Visual Results}
\begin{figure}[h!]
  \centering
  \includegraphics[width=\columnwidth]{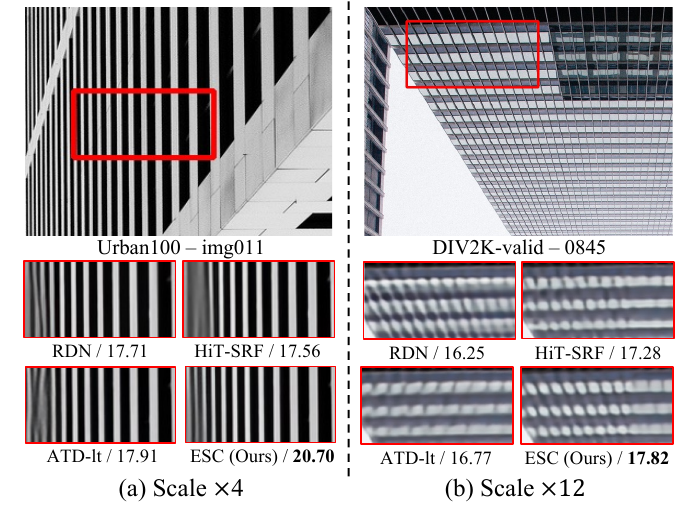}
  \vspace{-0.8cm}
  \caption{
    Visual comparisons on arbitrary-scale SR for both in-distribution~($\times$4) and out-of-distribution~($\times$12) scales.
    The best result on PSNR is bolded.
  }
  \vspace{-0.3cm}
  \label{fig:arb_visual}
\end{figure}

We also visually compare our network with RDN, HiT-SRF, and ATD-light.
As shown in Figure~\ref{fig:arb_visual}, Our network effectively restores fine details that other methods fail on both $\times$4 and $\times$12 scales.
This result demonstrates that our network also excels at restoring visually present results on both in-distribution and out-of-distribution scales.

\subsection{Real-World Super-Resolution}
\begin{figure}[t]
  \centering
  \includegraphics[width=\columnwidth]{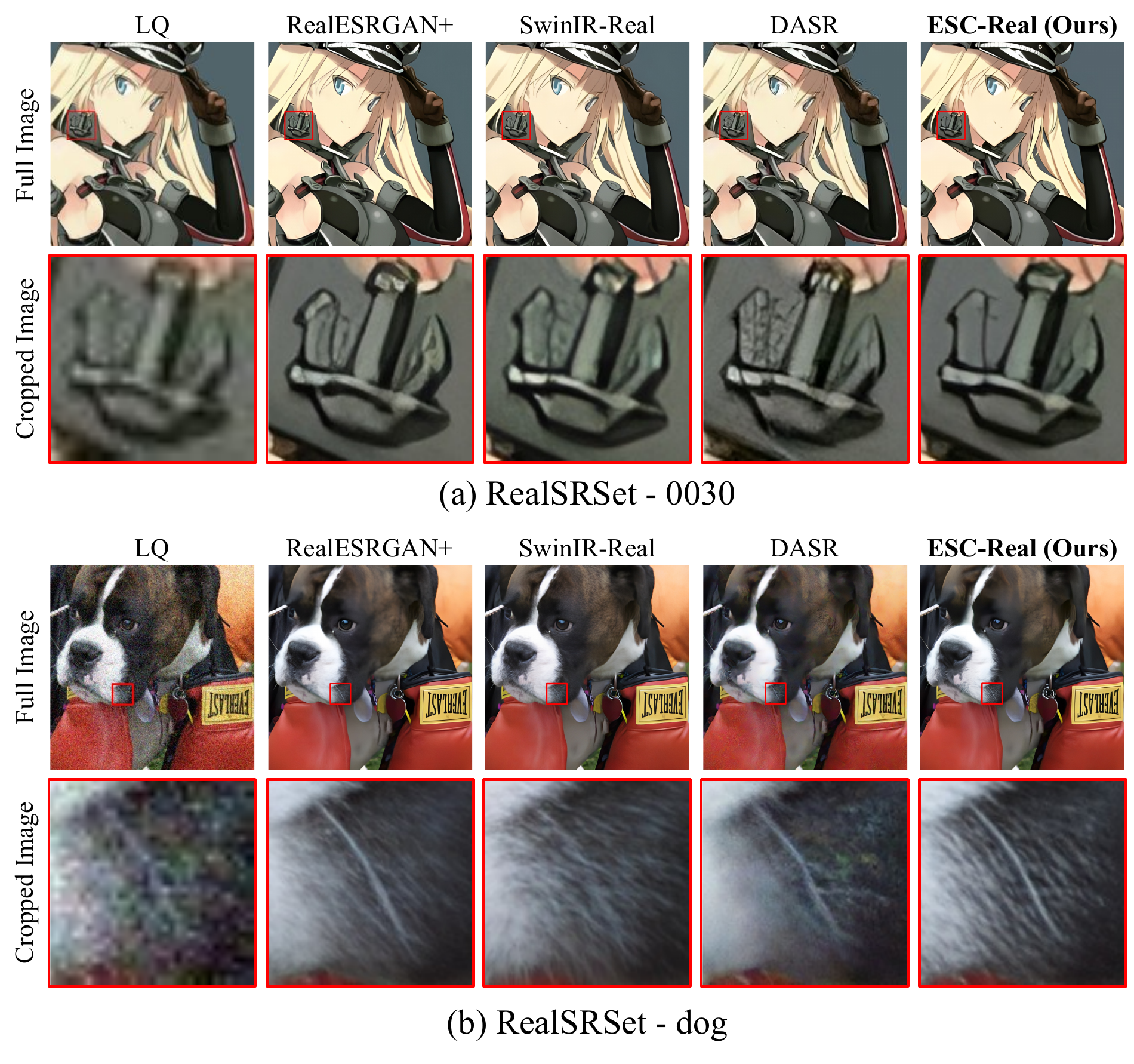}
  \caption{
    Visual comparisons on real-world SR~($\times$4) tasks.
  }
  \vspace{-0.5cm}
  \label{fig:real_visual}
\end{figure}

We further test our network on real-world SR tasks that aim to restore images with unknown degradations.
To this end, we introduce a variant called ESC-Real and compare them to representative real-world SR methods, including RealESRGAN+~\cite{RealESRGAN}, SwinIR-Real~\cite{SwinIR}, and DASR~\cite{DASR}.
For our evaluation, we use the commonly employed RealSRSet dataset~\cite{BSRGAN}.
As shown in Table~\ref{fig:real_visual}, our ESC-Real recovers more fine-grained details than the other methods while significantly reducing unpleasant artifacts in both graphic and natural images.
This result demonstrates that our proposed methods are effective even when dealing with complex unknown degradations, underscoring our methods' outstanding representational capability. 
Quantitative results on real-world SR tasks are presented in Section~\ref{sec:quant_rwsr} in the supplementary.

\vspace{-0.15cm}
\section{Conclusion}
In this paper, we propose ESC, a novel lightweight SR network designed to mitigate the excessive memory overhead of Transformers.
Our extensive experiments demonstrate that most self-attention layers can be effectively replaced with carefully designed convolutions to improve efficiency.
Moreover, we introduce flash attention into lightweight SR tasks, significantly enhancing performance without complicated modifications to self-attention.
As a result, we fully exploit the advantages of Transformers by prioritizing improvements in their efficiency.

\noindent \textbf{Acknowledgments}
This work was supported by the National Research Foundation (NRF) grant funded by the Korea government (MSIT) [RS-2025-00562400].

{
    \small
    \bibliographystyle{ieeenat_fullname}
    \bibliography{main}
}

\clearpage
\setcounter{page}{1}
\maketitlesupplementary

The supplementary includes implementation details of Flash Attention~(FA) and proposed networks, additional results on the arbitrary-scale super-resolution~(SR) tasks, quantitative results on real-world SR tasks, efficiency comparison beyond GPUs, and classic SR results on larger model sizes.

\section{Flash Attention Implementation Details}\label{sec:fa_imp_det}
FA~\cite{FlashAttn} implements self-attention by fusing kernels and avoiding materializing a full score matrix~($S=QK^{T}$), thereby significantly reducing both memory footprint and latency.
Although FA has been widely adopted across various domains, such as classification and generation, its application to SR tasks has been limited.
In order to alleviate the self‐attention's memory bottleneck, we attempted to integrate FA directly into SR architectures. 
However, we observed that naively applying FA leads to highly unstable training. 
We verified that this instability stems from FA’s fused‐kernel design, which blocks the use of relative positional bias~(RelPos)~\cite{T5, SwinTransformer}. 
As shown in Table~\ref{Tab:FA}, training loss diverges when FA is employed without RelPos.
A common alternative is to use Rotary Positional Encoding~(RoPE)~\cite{RoPE, RoPEViT}, which rotates query and key tensors for positional conditioning and therefore remains compatible with fused kernels~\cite{LightningDiT, CogVideoX}.
Nonetheless, our experiments demonstrate that even this approach fails to deliver acceptable performance in the SR setting.
To address these challenges, we leverage Flex Attention~\cite{FlexAttn}, which supports both user‐defined score modifications and FA. 
Our implementation not only resolves the memory bottleneck but also achieves superior performance by leveraging the 32$\times$32 window size for self-attention.

\begin{table}[h]
\caption{Comparisons on performance by FA implementation.}\label{Tab:FA}
\centering
\resizebox{\columnwidth}{!}{%
\begin{tabular}{@{}lccc@{}}
\toprule
Type& FA~\cite{FlashAttn} & FA w/ RoPE~\cite{RoPEViT} & \textbf{FA w/ RelPos~(Ours)} \\ \midrule
PSNR/SSRM~(U100$\times$2) & NaN / NaN & 32.76 / 0.9343 & \textbf{33.46} / \textbf{0.9395} \\ \bottomrule
\end{tabular}%
}
\end{table}

\section{Network Implementation Details}\label{sec:implementation_details}
This section describes the details of the implementation of our methods.
We begin by outlining the basic model configuration for each task, which includes the hyperparameters $C$, $N$, and $M$.
Here, $C$ represents the number of channels, $N$ denotes the number of $\mathrm{ESCBlock}$s, and $M$ indicates the number of $\mathrm{ConvAttn}$s. 
Next, we describe additional components, such as the compositions of $U$ and $S$, among others.
Finally, we present the training details, which cover the training datasets, the number of training iterations, the learning rate, optimizer configurations, the loss function used, and the training batch and patch size.

\subsection{Classic SR}
Three variants are introduced in the classic SR task: ESC-FP, ESC-light, and ESC. 
ESC-FP is a variant needed when it is necessary to reduce FLOPs and parameter size, while ESC-light and ESC are variants needed when it is necessary to reduce latency. 
The basic configurations for each variant are as follows: for ESC-FP, $C$, $N$, and $M$ are 48, 5, and 5; for ESC-light, they are 64, 3, and 5; and for ESC, they are 64, 5, and 5.
All variants use Sub-Pixel Convolution~(SPConv)~\cite{ESPCN} as $U$.
ESC-light and ESC utilize the repeat function~\cite{ABPN} as $S$ and add $F^{s}$ before the pixel shuffle of SPConv, while ESC-FP employs bicubic interpolation as $S$ and adds $F^{s}$ to the output of SPConv. 
Furthermore, ESC-FP employs decomposed $LK$ to further reduce FLOPs and parameter size and utilizes extra $LN$s in front of the $\mathrm{ConvFFN}$s.
Lastly, hidden dimension~($h$) for kernel estimators is set to 8 for ESC and ESC-FP, while ESC-FP uses 4.
For training, we use the DIV2K~\cite{DIV2K} dataset, and for the data scaling experiments, we employ the DIV2K+Flickr2K+LSDIR+Diverseg-IP~(DFLIP)~\cite{DF2KDataset, LSDIR, DiverSeg} dataset. 
Training our networks lasts for 500K iterations, and the optimizer used is AdamW~\cite{AdamW} with $\beta$s of 0.9 and 0.9 and a learning rate of 5e-4. 
We use L1 loss and 64 patches of size 64$\times$64 as input to train.
The networks of scale $\times3$ and $\times$4 are fine-tuned from the results of the $\times$2 scale.
To train other methods~\cite{SRFormer, HiTSR, ATD} for data scaling experiments, we follow the training details described in their paper.

\subsection{Arbitrary-scale SR}
For the Arbitrary-scale SR task, we use the same basic configuration as the ESC. 
The difference between the ESC in Classic SR and the ESC in arbitrary-scale SR is that LTE~\cite{LTE} is used as $U$, and accordingly, $S$ is also changed to bilinear interpolation.
Training details, including other models~\cite{ATD, HiTSR}, are the same as LTE across all instances, using the DIV2K dataset, running for 1000 epochs, utilizing the Adam~\cite{Adam} with $\beta$s values of 0.9 and 0.999, and leveraging L1 loss.
However, since HiT-SRF and our ESC are optimized for training with the input patches of 64$\times$64, we train all Transformers leveraging 32 input patches of size 64$\times$64.
Still, the number of sampling coordinates for training remains the same as RDN+LTE, which is 2304~($48^{2}$).

\begin{table*}[!ht]
\caption{
    Quantitative comparison on arbitrary-scale SR task employing LTE~\cite{LTE} as upsampler. The best result on PSNR is bolded.
}\label{tab:fixedscale_arb}
\resizebox{\textwidth}{!}{%
\begin{tabular}{@{}l|ccccc|ccccc|ccccc|ccccc@{}}
\toprule
\multirow{3}{*}{Methods} & \multicolumn{5}{c|}{Set5} & \multicolumn{5}{c|}{Set14} & \multicolumn{5}{c|}{B100} & \multicolumn{5}{c}{Urban100} \\ \cmidrule(l){2-21} 
 & \multicolumn{3}{c|}{Seen} & \multicolumn{2}{c|}{Unseen} & \multicolumn{3}{c|}{Seen} & \multicolumn{2}{c|}{Unseen} & \multicolumn{3}{c|}{Seen} & \multicolumn{2}{c|}{Unseen} & \multicolumn{3}{c|}{Seen} & \multicolumn{2}{c}{Unseen} \\ \cmidrule(l){2-21} 
 & $\times$2 & $\times$3 & \multicolumn{1}{c|}{$\times$4} & $\times$6 & $\times$8 & $\times$2 & $\times$3 & \multicolumn{1}{c|}{$\times$4} & $\times$6 & $\times$8 & $\times$2 & $\times$3 & \multicolumn{1}{c|}{$\times$4} & $\times$6 & $\times$8 & $\times$2 & $\times$3 & \multicolumn{1}{c|}{$\times$4} & $\times$6 & $\times$8 \\ \midrule
RDN+LTE~\cite{RDN} & 38.23 & 34.72 & \multicolumn{1}{c|}{32.61} & \textbf{29.32} & \textbf{27.26} & \textbf{34.09} & 30.58 & \multicolumn{1}{c|}{28.88} & \textbf{26.71} & 25.16 & 32.36 & 29.30 & \multicolumn{1}{c|}{\textbf{27.77}} & \textbf{26.01} & 24.95 & 33.04 & 28.97 & \multicolumn{1}{c|}{26.81} & 24.28 & 22.88 \\
ATD-lt+LTE$^{\S}$~\cite{ATD} & 38.28 & 34.73 & \multicolumn{1}{c|}{32.57} & 29.21 & \multicolumn{1}{c|}{27.22} & 34.14 & 30.64 & \multicolumn{1}{c|}{28.91} & 26.67 & \multicolumn{1}{c|}{25.21} & 32.35 & 29.30 & \multicolumn{1}{c|}{\textbf{27.77}} & \textbf{26.01} & \multicolumn{1}{c|}{24.95} & 33.12 & 29.06 & \multicolumn{1}{c|}{26.95} & 24.41 & 23.00 \\
HiT-SRF+LTE$^{\S}$~\cite{HiTSR} & 38.27 & 34.74 & \multicolumn{1}{c|}{32.59} & 29.25 & 27.21 & 34.03 & 30.64 & \multicolumn{1}{c|}{28.91} & 26.68 & 25.21 & 32.37 & 29.30 & \multicolumn{1}{c|}{27.76} & 26.00 & 24.94 & 33.18 & 29.05 & \multicolumn{1}{c|}{26.89} & 24.34 & 22.94 \\
\textbf{ESC+LTE~(Ours)} & \textbf{38.29} & \textbf{34.79} & \multicolumn{1}{c|}{\textbf{32.72}} & 29.18 & 27.24 & 34.05 & \textbf{30.69} & \multicolumn{1}{c|}{\textbf{28.94}} & 26.70 & \textbf{25.24} & \textbf{32.38} & \textbf{29.32} & \multicolumn{1}{c|}{\textbf{27.77}} & 26.00 & \textbf{24.96} & \textbf{33.30} & \textbf{29.21} & \multicolumn{1}{c|}{\textbf{27.04}} & \textbf{24.44} & \textbf{23.03} \\ \bottomrule
\end{tabular}%
}
\end{table*}

\subsection{Real-world SR}
For the real-world SR task, we introduce ESC-Real with a basic configuration of 64, 10, and 5, which denote $C$, $N$, and $M$, respectively.
ESC-Real employs the same upsampler as RealESRGAN~\cite{RealESRGAN} and SwinIR-Real~\cite{SwinIR}, utilizing it as $U$, and incorporates four layers of shallow network~(\texttt{c128k1g1}--\texttt{c128k7g128}--\texttt{LeakyReLU($\alpha=0.2$)}--\texttt{c64k1g1}) as $S$.
Here, $\texttt{c}$, $\texttt{k}$, and $\texttt{g}$ denote channel, kernel, and group size for convolution, respectively.
In this approach, $F^{s}$ is added into $F$.
We use the RealESRGAN degradation model and DF2KOST~\cite{STFGAN} dataset to generate low-quality images.
ESC-real is first trained for 1M iterations using L1 loss, then trained for 400K iterations with L1 loss, adversarial loss, and perceptual loss, using weight factors of 1, 0.1, and 1, respectively. 
The network architectures used for calculating adversarial loss and perceptual loss are the same as those used in RealESRGAN.
In both phases, 48 patches of size 64$\times$64 are used as input for training.

\section{Additional Results on Arbitrary-scale SR}\label{sec:add_res_assr}
This section exhibits additional results on the arbitrary-scale SR task.
Additional quantitative results are measures on the four commonly used evaluation datasets, including Set5~\cite{Set5}, Set14~\cite{Set14}, B100~\cite{BSD100}, and Urban100~\cite{Urban100}.
Following previous research~\cite{LTE}, we measure Peak Signal-to-Noise Ratio~(PSNR) on the Y channel after cropping the image's boundary equivalent to the upscaling factor and converting it to YCbCr color space.
Our ESC+LTE outperforms other methods on Urban100 at both seen and unseen scales, as shown in Table~\ref{tab:fixedscale_arb}, 

\begin{table}[!ht]
\caption{
    Quantitative comparisons on real-world SR.
}\label{tab:real}
\vspace{-0.2cm}
\resizebox{\columnwidth}{!}{%
\begin{tabular}{@{}l|l|cccc@{}}
\toprule
Dataset & Metrics & RealESRGAN+ & SwinIR-Real & DASR & \textbf{ESC-Real} \\ \midrule
\multirow{4}{*}{RealLQ250~\cite{DreamClear}} & NIQE $\downarrow$ & 4.1328 & 4.1779 & 4.7858 & 4.0556 \\
 & MANIQA $\uparrow$ & 0.3564 & 0.3400 & 0.2789 & 0.3553 \\
 & MUSIQ $\uparrow$ & 62.51 & 60.48 & 53.02 & 62.98 \\
 & CLIPIQA $\uparrow$ & 0.5437 & 0.5348 & 0.4631 & 0.5796 \\ \midrule
\multirow{4}{*}{RealSRSet~\cite{BSRGAN}} & NIQE $\downarrow$ & 5.3430 & 5.1037 & 4.5931 & 5.0181 \\
 & MANIQA $\uparrow$ & 0.3988 & 0.3872 & 0.3277 & 0.3952 \\
 & MUSIQ $\uparrow$ & 64.25 & 63.68 & 58.82 & 64.58 \\
 & CLIPIQA $\uparrow$ & 0.5942 & 0.5921 & 0.5278 & 0.6156 \\ \midrule
\multirow{9}{*}{RealSR~\cite{RealSR}} & PSNR $\uparrow$ & 24.53 & 24.71 & 25.86 & 24.52 \\
 & SSIM $\uparrow$ & 0.7484 & 0.7547 & 0.7617 & 0.7503 \\
 & LPIPS $\downarrow$ & 0.2729 & 0.2594 & 0.3113 & 0.2622 \\
 & DISTS $\downarrow$ & 0.1685 & 0.1609 & 0.1838 & 0.1671 \\
 & FID $\downarrow$ & 67.01 & 64.19 & 63.62 & 66.81 \\
 & NIQE $\downarrow$ & 4.6801 & 4.6465 & 5.9682 & 4.4848 \\
 & MANIQA $\uparrow$ & 0.3675 & 0.3504 & 0.2663 & 0.3799 \\
 & MUSIQ $\uparrow$ & 59.69 & 59.64 & 45.82 & 61.40 \\
 & CLIPIQA $\uparrow$ & 0.4903 & 0.4736 & 0.3629 & 0.5338 \\ \midrule
\multirow{9}{*}{DRealSR~\cite{DRealSR}} & PSNR $\uparrow$ & 26.59 & 26.52 & 28.40 & 26.76 \\
 & SSIM $\uparrow$ & 0.7988 & 0.7923 & 0.8302 & 0.79534 \\
 & LPIPS $\downarrow$ & 0.2818 & 0.2838 & 0.2962 & 0.2795 \\
 & DISTS $\downarrow$ & 0.1464 & 0.1461 & 0.1689 & 0.1503 \\
 & FID $\downarrow$ & 23.19 & 24.63 & 17.89 & 21.35 \\
 & NIQE $\downarrow$ & 4.7164 & 4.5683 & 6.3473 & 4.7006 \\
 & MANIQA $\uparrow$ & 0.3431 & 0.3275 & 0.2733 & 0.3442 \\
 & MUSIQ $\uparrow$ & 35.27 & 34.62 & 28.63 & 35.21 \\
 & CLIPIQA $\uparrow$ & 0.5179 & 0.5039 & 0.3843 & 0.5564 \\ \bottomrule
\end{tabular}%
}
\end{table}

\section{Quantitative Results on Real-world SR}\label{sec:quant_rwsr}
Taking a step further, we report quantitative results for the real-world SR task. 
To this end, we measure a variety of metrics~(PSNR@Y, SSIM, LPIPS~\cite{LPIPS}, DISTS~\cite{DISTS}, FID~\cite{FID}, NIQE~\cite{NIQE}, MANIQA~\cite{MANIQA}, MUSIQ~\cite{MUSIQ}, and CLIP-IQA~\cite{CLIPIQA}) on multiple datasets (RealLQ250~\cite{DreamClear}, RealSRSet~\cite{BSRGAN}, RealSR~\cite{RealSR}, and DRealSR~\cite{DRealSR}).
As shown in Table \ref{tab:real}, ESC-Real achieves the highest CLIPIQA scores on all datasets, demonstrating its ability to produce reconstructions with superior perceptual quality.

\begin{table}[h]
\caption{Comparisons of latency on MacBook M2 air, and iPhone 12}\label{Tab:Lat}
\renewcommand{\arraystretch}{0.95}
\centering
\resizebox{\columnwidth}{!}{
\begin{tabular}{@{}lcccccc@{}}
\toprule
Methods & ELAN-lt & OmniSR & ASID-D8 & HiT-SRF & \textbf{ESC-lt} & \textbf{ESC} \\ \midrule
M2Air~($X\in\mathbb{R}^{128\times128\times3}$) & 318.51 & Failed & Failed & 88145.85 & \textbf{124.07} & \textbf{181.16} \\ 
iPhone12~($X\in\mathbb{R}^{32\times32\times3}$) & 38.12 & Failed & Failed & OOM & \textbf{25.57} & \textbf{42.78} \\ 
\bottomrule
\end{tabular}%
}
\end{table}

\section{Efficiency Comparisons beyond GPUs}\label{sec:add_latency}
In real‐world deployment scenarios, networks often run on devices with far more constrained resources than GPUs. 
To evaluate our method under such conditions, we benchmarked several Transformer‐based SR models~(ELAN~\cite{ELAN}, OmniSR~\cite{OmniSR}, ASID‐D8~\cite{ASID}, HiT‐SRF~\cite{HiTSR}) against our ESC(‐lt) on a MacBook Air M2 and an iPhone 12. 
As detailed in Table~\ref{Tab:Lat}, whereas the other Transformers either fail to compile or incur out-of-memory errors, ESC‐lt achieves up to a 61\% reduction in latency compared to ELAN-light, demonstrating its efficiency in real‐world deployments.

\begin{table*}[!ht]
\caption{
    Comparisons of larger classic SR methods~(Params$>$10M). PT denotes pre-training with 64$\times$64 patches and FT denotes fine-tuning with 96$\times$96 patches.
}\label{tab:fixedscale_largermodel}
\resizebox{\textwidth}{!}{%
\begin{tabular}{@{}l|c|c|ccccc@{}}
\toprule
\multirow{2}{*}{Method} & \multirow{2}{*}{Scale} & \multirow{2}{*}{\#Params~(M)} & \multicolumn{5}{c}{PSNR / SSIM} \\
 &  &  & Set5 & Set14 & B100 & Urban100 & Manga109 \\ \midrule
SwinIR~\cite{SwinIR} & \multirow{8}{*}{$\times2$} & 11.8 & 38.42/0.9623 & 34.46/0.9250 & 32.53/0.9041 & 33.81/0.9433 & 39.92/0.9797 \\
EDT-B~\cite{EDT} &  & 11.5 & 38.63/0.9632 & 34.80/0.9273 & 32.62/0.9052 & 34.27/0.9456 & 40.37/0.9811 \\
CAT-A~\cite{CAT} &  & 16.5 & 38.51/0.9626 & 34.78/0.9265 & 32.59/0.9047 & 34.26/0.9440 & 40.10/0.9805 \\
ART~\cite{ART} &  & 16.4 & 38.56/0.9629 & 34.59/0.9267 & 32.58/0.9048 & 34.30/0.9452 & 40.24/0.9808 \\
ACT~\cite{ACT} &  & 46.0 & 38.46/0.9626 & 34.60/0.9256 & 32.56/0.9048 & 34.07/0.9443 & 39.95/0.9804 \\
SRFormer~\cite{SRFormer} &  & 10.5 & 38.51/0.9627 & 34.44/0.9253 & 32.57/0.9046 & 34.09/0.9449 & 40.07/0.9802 \\
\textbf{ESC~(PT)} &  & 12.5 & 38.52/0.9626 & 34.57/0.9257 & 32.58/0.9045 & 34.24/0.9450 & 40.18/0.9803 \\
\textbf{ESC~(FT)} &  & 12.5 & 38.59/0.9630 & 34.70/0.9259 & 32.61/0.9052 & 34.49/0.9466 & 40.38/0.9809 \\ \midrule
SwinIR~\cite{SwinIR} & \multirow{7}{*}{$\times3$} & 11.9 & 34.97/0.9318 & 30.93/0.8534 & 29.46/0.8145 & 29.75/0.8826 & 35.12/0.9537 \\
EDT-B~\cite{EDT} &  & 11.7 & 35.13/0.9328 & 31.09/0.8553 & 29.53/0.8165 & 30.07/0.8863 & 35.47/0.9550 \\
CAT-A~\cite{CAT} &  & 16.6 & 35.06/0.9326 & 31.04/0.8538 & 29.52/0.8160 & 30.12/0.8862 & 35.38/0.9546 \\
ART~\cite{ART} &  & 16.6 & 35.07/0.9325 & 31.02/0.8541 & 29.51/0.8159 & 30.10/0.8871 & 35.39/0.9548 \\
ACT~\cite{ACT} &  & 46.0 & 35.03/0.9321 & 31.08/0.8541 & 29.51/0.8164 & 30.08/0.8858 & 35.27/0.9540 \\
SRFormer~\cite{SRFormer} &  & 10.7 & 35.02/0.9323 & 30.94/0.8540 & 29.48/0.8156 & 30.04/0.8865 & 35.26/0.9543 \\
\textbf{ESC~(FT)} &  & 12.5 & 35.14/0.9330 & 31.10/0.8552 & 29.53/0.8167 & 30.23/0.8895 & 35.60/0.9555 \\ \midrule
SwinIR~\cite{SwinIR} & \multirow{7}{*}{$\times4$} & 11.9 & 32.92/0.9044 & 29.09/0.7950 & 27.92/0.7489 & 27.45/0.8254 & 32.03/0.9260 \\
EDT-B~\cite{EDT} &  & 11.6 & 33.06/0.9055 & 29.23/0.7971 & 27.99/0.7510 & 27.75/0.8317 & 32.39/0.9283 \\
CAT-A~\cite{CAT} &  & 16.6 & 33.08/0.9052 & 29.18/0.7960 & 27.99/0.7510 & 27.89/0.8339 & 32.39/0.9285 \\
ART~\cite{ART} &  & 16.6 & 33.04/0.9051 & 29.16/0.7958 & 27.97/0.7510 & 27.77/0.8321 & 32.31/0.9283 \\
ACT~\cite{ACT} &  & 46.0 & 32.97/0.9031 & 29.18/0.7954 & 27.95/0.7507 & 27.74/0.8305 & 32.20/0.9267 \\
SRFormer~\cite{SRFormer} &  & 10.6 & 32.93/0.9041 & 29.08/0.7953 & 27.94/0.7502 & 27.68/0.8311 & 32.21/0.9271 \\
\textbf{ESC~(FT)} &  & 12.5 & 33.00/0.9054 & 29.21/0.7968 & 27.95/0.7504 & 27.89/0.8351 & 32.54/0.9295 \\ \bottomrule
\end{tabular}%
}
\end{table*}

\section{Classic SR Results on Larger Model Size}\label{sec:larger_model}
Although we have demonstrated substantial performance gains over lightweight Transformers~(Params$<$1M), larger models~(Params$>$10M) remain an active area of research. 
To assess our method in this regime, we scale ESC to 12.5M parameters, on par with the size of SwinIR, and train and evaluate it accordingly. 
The scaled ESC uses the window size of 48$\times$48, $N=8$, $M=5$, $C=192$, $C_{\mathrm{ConvAttn}}=48$, and $h=24$.
Extra layer normalizations are placed before the $\mathrm{ConvFFN}$s.
We leverage the DF2K dataset and follow the ATD's training strategy~\cite{ATD}, pre-training on small patches (64$\times$64) and then fine-tuning on larger patches (96$\times$96).
Table~\ref{tab:fixedscale_largermodel} shows that ESC delivers performance on par with other large-scale SR Transformers.

\end{document}